\documentclass[11pt]{article}
\ifdefined\XeTeXrevision\else\ifdefined\pdfoutput\pdfoutput=1\fi\fi

\usepackage[final]{acl}

\usepackage{times}
\usepackage{latexsym}

\usepackage[T1]{fontenc}

\usepackage[utf8]{inputenc}

\usepackage{microtype}

\usepackage{inconsolata}

\usepackage{graphicx}

\usepackage[colorinlistoftodos, textsize=small]{todonotes}

\usepackage{booktabs} %
\usepackage{multirow} %
\usepackage{cuted} %
\usepackage{amsmath} %
\usepackage{amssymb} %
\usepackage{bm} %
\usepackage{tikz}
\usepackage{colortbl}
\usepackage{enumitem}

\title{Zero-Shot Cross-Lingual Recognition of Sign Language Handshapes}

\author{
 \textbf{Marcel Granero-Moya\textsuperscript{*}},
 \textbf{Carolina del Corral Farrarós\textsuperscript{*}},
 \\
 \textbf{Gloria Haro},
 \textbf{Coloma Ballester},
 \textbf{Ricardo Marques}
\\
\\
 Universitat Pompeu Fabra
}

\begin{document}

\maketitle

\begingroup
\renewcommand\thefootnote{}%
\footnotetext{\textsuperscript{*}\,Equal contribution.\\
Correspondence: \href{mailto:marcel.granero@upf.edu}{marcel.granero@upf.edu}}
\endgroup

\begin{abstract}
Sign language processing advances rapidly for high-resource languages such as American Sign Language (ASL), yet most of the world's sign languages lack the phonological annotations new methods require. We present the first zero-shot cross-lingual framework for handshape recognition, transferring from ASL to Catalan Sign Language (LSC). Our approach leverages the decomposition of handshapes into five phonological features — selected fingers, flexion, spread, thumb position, and thumb contact — shared across both languages, to decode LSC handshapes from predicted features via a composite phonological distance metric. We evaluate three architectures (MLP, SL-GCN, SHuBERT) trained on two ASL corpora (PopSign, Sem-Lex) against a 37-handshape, single-signer LSC benchmark. Zero-shot transfer proves viable once recording-format disparities are harmonized, reaching 80.0\% phonological feature accuracy and 54.5\% expected handshape accuracy. Phonological decomposition thus offers a bridge for extending sign language technologies to low-resource languages without any target-language video training labels.
\end{abstract}

\section{Introduction}
While over 70 million deaf people globally use over 200 distinct sign languages (SLs) \cite{WFD_stats}, most remain low-resource \cite{bragg2019sign}. Annotated corpora and modeling efforts are heavily concentrated on a few high-resource languages like American Sign Language (ASL). In contrast, Catalan Sign Language (LSC) lacks the annotated resources needed to drive modern recognition systems, beyond a gloss-annotated dataset \cite{iec2025corpuslsc} and comprehensive grammar \cite{quer2020grammar}.

Sign Language Processing (SLP) is bottlenecked by annotation costs: labels are costly and require fluent experts, and gloss-level supervision does not transfer across distinct lexicons. Phonological features offer a compelling solution. They provide a linguistically grounded label set that bridges languages. Because handshapes decompose into core phonological features---\textit{selected fingers}, \textit{finger flexion}, \textit{spread}, \textit{thumb position}, and \textit{thumb contact} \cite{brentari1998prosodic}---a model trained on a high-resource language could theoretically recognize unseen target-language handshapes by recombining these shared phonological features.

While monolingual phonological recognition is well established \cite{kezar2023sem, Handshape-GNN, shubert, inoue-etal-2026-resource}, and prior cross-lingual modeling relies on target-language annotations \cite{tornay2020towards, bilge2024crosslingual}, strictly zero-shot cross-lingual phonological recognition remains unexplored.

We address this gap by transferring from ASL to LSC. Our contributions are as follows.
\begin{itemize}[nolistsep]
\item The first zero-shot cross-lingual handshape recognition framework, evaluated across three architectural paradigms (static MLP, SL-GCN \cite{jiang2021skeleton}, and SHuBERT \cite{shubert}), using two ASL corpora (PopSign \cite{starner2023popsign} and Sem-Lex \cite{kezar2023sem}).
\item A formal phonological alignment scheme mapping \citeposs{navarrete2020phonology} LSC handshapes into the ASL-LEX 2.0 \cite{sehyr2021asl} phonological feature space.
\item A distance-based decoding protocol and expected-accuracy metric designed to handle mappings from phonological features to target-language handshapes.
\item Empirical evidence that transfer is viable---achieving up to 80.0\% phonological feature accuracy and 54.5\% expected handshape accuracy---but only after harmonizing recording-format disparities.
\end{itemize}

\section{Related work}

Our contribution bridges five lines of work that structure this section. Cross-lingual phonological transfer relies on three foundations: a shared feature inventory (\emph{phonology of sign languages}), robust sign encoding models (\emph{sign language processing}), and annotated resources (\emph{phonology-curated datasets}). While their convergence enables monolingual \emph{phonology recognition}, extending this to \emph{cross-lingual phonological recognition} remains an open frontier—where we position this work.

\paragraph{Phonology of Sign Languages}
\citet{stokoe1960sign} and \citet{battison1978lexical} established that Manual Features of signs decompose into contrastive sub-lexical units: handshape, palm orientation, location, and movement. Subsequent frameworks like \citeposs{brentari1998prosodic} Prosodic Model refined these into a hierarchical feature geometry, decomposing handshapes into finer-grained features such as selected fingers, finger flexion, spread, and thumb position; and decomposing movement into path and location components. Modern datasets like ASL-LEX \cite{caselli2017asl, sehyr2021asl} code phonology precisely at this granular level.
Crucially, LSC relies on these same sub-handshape parameters \cite{quer2020grammar, navarrete2020phonology}. This structural overlap, stemming from their shared lineage in the \textit{Francosign} family \cite{wittmann1991classification}, motivates the cross-linguistic transfer pursued in this work.

\paragraph{Sign Language Processing}
Early SLP relied on gloss-based pipelines constrained by comparatively small annotated datasets like RWTH-PHOENIX-Weather dataset \cite{forster2012rwth, forster2014extensions, camgoz2018neural, camgoz2020sign}. Recent advances have shifted the paradigm toward gloss-free translation \cite{zhou2023gloss, lin2023gloss, muller2023considerations}, mapping video or pose sequences directly to text, but critically depend on robust visual features. To address this, recent work leverages self-supervised pre-training on large, unannotated video corpora like YouTube-ASL \cite{uthus2023youtube} and YouTube-SL-25 \cite{tanzer2025youtube}. Models like SHuBERT \cite{shubert} use these datasets to learn generalized spatio-temporal representations without costly gloss annotations.

\paragraph{Phonology-Curated Datasets}
Interpreting fine-grained sub-lexical structures requires dedicated phonological resources. Several curated datasets have emerged, though available resources remain overwhelmingly restricted to ASL: ASL-LEX 2.0 \cite{sehyr2021asl} catalogs properties for over 2,700 glosses; Sem-Lex \cite{kezar2023sem} aligns deaf signers' isolated productions to ASL-LEX 2.0; and PopSign \cite{starner2023popsign, chow2023google} offers game-collected sign recordings linked to these phonological features. We refer the reader to Section~\ref{sec:datasets} for further dataset details.

\paragraph{Phonology Recognition}
Early work focused on isolated feature extraction, with milestones like DeepHand \cite{koller2016deep} targeting handshape recognition. To enable deeper phonological understanding, subsequent systems have leveraged on the aforementioned curated datasets to capture complex spatial and temporal variations.
For example, \citet{kezar2023sem} demonstrated that phonology recognition is an effective auxiliary target for Isolated Sign Language Recognition (ISLR), showing that an SL-GCN \cite{jiang2021skeleton} trained on Sem-Lex achieves 85\% average accuracy across the 16 ASL-LEX phonological feature types.
For handshape recognition, the Handshape-GNN model \cite{Handshape-GNN}---trained for the PopSign Kaggle challenge \cite{chow2023google}---separates temporal sign dynamics from static hand configurations, benchmarking effectively against a multilayer perceptron (MLP) baseline.
Similarly, the self-supervised SHuBERT \cite{shubert} achieved state-of-the-art ISLR accuracy on Sem-Lex when fine-tuned, suggesting that its learned representations implicitly encode sub-lexical phonological structure.

\paragraph{Cross-lingual Phonological Recognition}
Despite successes with rich, language-specific training data, the aforementioned models have not been evaluated on lower resourced SLs in a cross-lingual setup.
Prior cross-lingual approaches adapt phonological subunits on target-language data \cite{tornay2020towards} or recognize novel signs from a few labeled target-language examples \cite{bilge2024crosslingual}. However, target-language annotation is a prohibitively expensive bottleneck for under-documented sign languages. A zero-shot approach bypasses this by leveraging shared cross-linguistic phonological structures. To our knowledge, zero-shot cross-lingual sign phonology prediction remains unexplored. We address this gap, demonstrating successful zero-shot transfer and establishing the first baseline for this task.

\section{Datasets}
\label{sec:datasets}
To evaluate cross-lingual transfer across diverse capture conditions and signers, our framework combines three ASL resources with a target LSC benchmark. Table~\ref{tab:dataset_summary} synthesizes their primary characteristics.

\textbf{ASL-LEX 2.0} \cite{sehyr2021asl} serves as the ASL lexical-phonological reference dictionary, providing over 2,700 glosses with detailed phonological annotations. While the dataset tracks 16 phonological features, our handshape prediction pipeline employs the 5 phonological features strictly related to the handshape (\textit{selected fingers}, \textit{flexion}, \textit{spread}, \textit{thumb position}, and \textit{thumb contact}) plus handshape class. Crucially, the combination of these 5 features is not always a strict 1-to-1 mapping. As detailed by \citet{sehyr2021asl}, phonological feature combinations can group non-contrastive variants together or fail to differentiate contrastive handshapes that vary in unselected finger flexion (see Section ~\ref{par:phonological_representation_limits_and_extensions}).

\textbf{Sem-Lex} \cite{kezar2023sem} connects ASL-LEX 2.0 annotations to a large-scale corpus of over 84,000 webcam video instances representing 3,149 signs recorded remotely by 41 Deaf signers. The dataset combines ASL-LEX citation signs, free-text responses, and SignBank \cite{hochgesang2019signbank} entries. Due to reproducibility issues with the official pre-extracted poses, we re-extracted all poses directly from the raw video source.

\textbf{PopSign} \citep{starner2023popsign} introduces in-the-wild mobile recordings collected from 47 ASL learners, comprising ${\sim}175\text{k}$ videos covering 250 isolated glosses balanced across all three dataset splits. We utilize the raw video release of the game split rather than the landmark-only Kaggle release \cite{chow2023google}, as RGB source videos are required for hybrid multimodal architectures (e.g., SHuBERT) that process pixel features alongside pose sequences.

\textbf{LSC Benchmark}\footnote{A pre-release subset was shared with us prior to publication; redistribution and visual sample display are restricted by the dataset creators.} was accessed as a pre-release subset to evaluate zero-shot cross-lingual transfer. It comprises 37 target handshapes performed by a native Deaf LSC signer. Structured according to \href{https://thesignhub.eu/grammar/lsc?tag=82}{SignHub's LSC finger configuration} \cite{navarrete2020phonology}, each handshape contains two videos: (1) a target handshape demonstration, and (2) an example sign.

\begin{table}[tb]
    \centering
    \renewcommand{\arraystretch}{1.2}
    \setlength{\tabcolsep}{4pt}
    \resizebox{\columnwidth}{!}{%
        \begin{tabular}{l l l l l l l}
            \toprule
            \textbf{Dataset} & \textbf{Lang.} & \textbf{\#Hs} & \textbf{\#Glosses} & \textbf{\#Videos} & \textbf{\#Signers} & \textbf{Rec. Type} \\ 
            \midrule
            \textbf{ASL-LEX 2.0}$^\dagger$ & ASL & 58 & 2,723 & 2,723 & -- & Studio \\ 
            \textbf{Sem-Lex} & ASL & 58 & 3,149 & 84,568 & 41 Deaf & Webcam \\ 
            \textbf{PopSign} & ASL & 37 & 250 & 175,022 & 47 Learners & Smartphone \\ 
            \textbf{LSC Benchmark} & LSC & 37 & -- & 74 & 1 Deaf & Studio \\ 
            \bottomrule
        \end{tabular}%
    }
    \caption{Overview and comparison of datasets evaluated in our experiments. Lang.: sign language; \#Hs.: number of unique handshape classes; Rec.\ Type: video recording environment or device. $^\dagger$ASL-LEX 2.0 was used strictly as a lexical mapping reference; its video data was not utilized in our experiments.}
    \label{tab:dataset_summary}
\end{table}

\section{Methodology}
\subsection{Model Architectures}
To benchmark phonological feature prediction and cross-lingual transferability, we evaluate three representative architectural paradigms: a lightweight pose-based MLP baseline, a spatio-temporal graph neural network (SL-GCN) \cite{jiang2021skeleton}, and a multi-modal pre-trained transformer (SHuBERT) \cite{shubert}. Each architecture is trained and evaluated across two distinct ASL source datasets spanning different capture modalities: Sem-Lex and PopSign.
To ensure complete reproducibility, we publicly release our model implementations, hyperparameter configurations, and evaluation pipelines, together with trained model checkpoints.\footnote{\href{https://marcelgranero.github.io/cross-lingual-handshapes}{https://marcelgranero.github.io/cross-lingual-handshapes}}

\subsubsection{Baseline MLP}
Following the static baseline architecture introduced by \citet{Handshape-GNN}, we employ a Multi-Layer Perceptron (MLP) with 3 layers of dimensions $63 \rightarrow 256 \rightarrow 256 \rightarrow N_{\text{classes}}$. The model operates strictly on static spatial features, removing the temporal dimension by extracting a single representative frame per sign sequence corresponding to minimal hand motion. Input features consist of raw 3D hand landmarks ($21 \times 3 = 63$ features) extracted via MediaPipe \cite{lugaresi2019mediapipe}. The network is trained from scratch on Sem-Lex and PopSign independently to predict multi-label phonological feature targets. This model serves as a computationally efficient reference to compare with more complex temporal and multi-modal architectures.

\subsubsection{SL-GCN}
To capture dynamic spatio-temporal hand movements and finger interactions, we employ the Sign Language Graph Convolutional Network (SL-GCN) \citep{jiang2021skeleton}, which served as the primary baseline for the Sem-Lex benchmark \cite{kezar2023sem}. SL-GCN models pose sequences as spatio-temporal graphs, applying graph convolutions across skeletal joints and temporal convolutions across consecutive frames. Following \citeposs{kezar2023sem}, we first train the network on ISLR before fine-tuning for joint gloss and phonological feature prediction across both Sem-Lex and PopSign. To maintain a unified extraction pipeline across both datasets (Section~\ref{sec:datasets}) and achieve optimal pose fidelity, all input pose sequences are re-extracted via MediaPipe \cite{lugaresi2019mediapipe} directly from the raw video sources. The complete reproduction benchmarks and validation metrics are provided in Table~\ref{tab:slgcn_table5_repro} (Appendix~\ref{app:reproduction_results}).

\subsubsection{SHuBERT}
To test whether large-scale video-pose pre-training enhances zero-shot cross-lingual transfer compared to purely pose-based models, we evaluate SHuBERT \cite{shubert}. SHuBERT is a self-supervised multimodal transformer pre-trained on YouTube-ASL \cite{uthus2023youtube} that fuses raw video pixels with pose sequences.
We fine-tune the pre-trained encoder for phonological feature prediction on Sem-Lex and PopSign. Reproduction performance benchmarks against the original paper are detailed in Table \ref{tab:shubert_table9_repro} (Appendix \ref{app:reproduction_results}).

\subsection{Zero-Shot Cross-Lingual Transfer}
To evaluate zero-shot handshape recognition across languages, we map target LSC handshapes to five phonological features in ASL-LEX format. Models trained on ASL predict this shared phonological feature space, and we map the resulting predicted features to target LSC handshape classes with a distance-based metric.
Note that this protocol is zero-shot with respect to target-language videos and video-level labels, but it does presuppose a dictionary-level target resource: a predefined LSC handshape inventory and its handshape-to-feature mapping (Appendix~\ref{sec:appendix-handshapes}).

\subsubsection{Aligning LSC Handshapes to ASL-LEX Phonological Features}
\label{subsec:mapping_lsc_handshapes_to_asl-lex}
To bridge the structural differences between ASL-LEX 2.0 and the \href{https://thesignhub.eu/grammar/lsc?tag=82}{SignHub LSC annotations} (SH-LSC), we establish a rule-based alignment that deterministically assigns the five ASL-LEX phonological features pertaining strictly to handshape configuration ---\textit{selected fingers}, \textit{finger flexion}, \textit{spread}, \textit{thumb position}, and \textit{thumb contact}--- to each LSC handshape
(detailed in Appendices \ref{sec:appendix-handshapes} and \ref{app:phonological_feature_inventories_across_datasets}). Specifically, Spread and Thumb Contact are mapped directly from SH-LSC to ASL-LEX categories without structural modifications. Finger flexion is mapped by harmonizing label conventions across schemes. For example, SH-LSC \textit{Extended} is mapped to ASL-LEX \textit{FullyOpen}. Additionally, because SH-LSC uses a single \textit{curved} category, \textit{curved} maps to a combined \textit{Curved/Bent} class formed by joining ASL-LEX \textit{Curved} and \textit{Bent}. Regarding thumb phonology, SH-LSC treats the thumb as a standard selected finger, whereas ASL-LEX excludes the thumb from \textit{Selected Fingers} unless it is the sole active digit. To align these paradigms, we map SH-LSC \textit{Selected Fingers} onto ASL-LEX \textit{Selected Fingers} and \textit{Thumb Position}: if an SH-LSC handshape includes the thumb $t$\footnote{Selected fingers are denoted as thumb $t$, index $i$, middle $m$, ring $r$, and pinkie $p$.} in its selected set, $t$ is re-encoded as $\textit{Thumb Position} = \textit{Open}$ and removed from \textit{Selected Fingers}, unless $t$ is the only selected digit (e.g., SH-LSC $\{t, i, m\} \rightarrow \text{ASL-LEX } \{i, m\} + \textit{Open}$).

\subsubsection{Phonological Handshape Matching and Evaluation}
\paragraph{Phonological Representation Space}
Let $\mathcal{P}_{\text{sel}}, \mathcal{P}_{\text{flex}}, \mathcal{P}_{\text{sprd}}, \mathcal{P}_{\text{th\_pos}}, \mathcal{P}_{\text{th\_cnt}}$ be discrete categorical sets representing five articulatory features: \textit{selected fingers}, \textit{finger flexion}, \textit{spread}, \textit{thumb position}, and \textit{thumb contact}, respectively. The phonological domain $\mathcal{P}$ is defined as their Cartesian product:
$\mathcal{P} = \mathcal{P}_{\text{sel}} \times \mathcal{P}_{\text{flex}} \times \mathcal{P}_{\text{sprd}} \times \mathcal{P}_{\text{th\_pos}} \times \mathcal{P}_{\text{th\_cnt}}$
An arbitrary phonological configuration is thus represented as a 5-dimensional column vector $\mathbf{p} = [p_{\text{sel}}, p_{\text{flex}}, p_{\text{sprd}}, p_{\text{th\_pos}}, p_{\text{th\_cnt}}]^T \in \mathcal{P}$, where each component $p_k$ takes a value from its corresponding feature set $\mathcal{P}_k$.
Let $\mathcal{H} = \{h_1, h_2, \dots, h_C\}$ denote the set of $C = 37$ target LSC handshape classes. Each class $h \in \mathcal{H}$ is associated with a canonical phonological feature vector $\mathbf{p}^{(h)} \in \mathcal{P}$ (See Table \ref{tab:handshape-inventory}). We define the set of valid LSC representations as
$\mathcal{P}_{\text{LSC}} = \left\{ \mathbf{p}^{(h)} \mid h \in \mathcal{H} \right\} \subset \mathcal{P}$.
Notably, the handshape-to-feature mapping $h \mapsto \mathbf{p}^{(h)}$ is non-injective:
two pairs\footnote{\texttt{7-flat\_open} with \texttt{index+thumb-flat\_open} and \texttt{7-flat\_closed} with \texttt{index+thumb-flat\_closed} coincide on phonological features. These handshapes differ on the unselected fingers flexion -- closed or open for \texttt{7} and \texttt{index+thumb} respectively.}
among the $37$ handshape classes share identical feature vectors, yielding $\vert{}\mathcal{P}_{\text{LSC}}\vert{} = 35$ unique canonical feature configurations.
Combinations in $\mathcal{P}$ that do not correspond to any valid LSC handshape form the set of unassociated phonological vectors $\mathcal{U} = \mathcal{P} \setminus \mathcal{P}_{\text{LSC}}$.

\paragraph{Inference and Constrained Decoding}
During inference, the ASL-trained models generate categorical predictions for the five phonological channels. 
Because ASL's \textit{selected fingers} and \textit{finger flexion} features contain more classes than those in LSC (see Appendix \ref{app:phonological_feature_inventories_across_datasets}),
we restrict each model's predictions to valid LSC target classes. We achieve this by applying a target-space mask to the output logits prior to the $\operatorname{argmax}$ operation\footnote{We mask three \textit{flexion} classes (\textit{Crossed}, \textit{Stacked}, and \textit{NA}) and four \textit{selected fingers} combinations (\textit{imp}, \textit{mr}, \textit{mrp}, and \textit{r}).}, yielding a predicted feature vector $\hat{\mathbf{p}} = [\hat{p}_{\text{sel}}, \hat{p}_{\text{flex}}, \hat{p}_{\text{sprd}}, \hat{p}_{\text{th\_pos}}, \hat{p}_{\text{th\_cnt}}]^T \in \mathcal{P}$. 
However, since each feature is predicted by an independent classification head, nothing prevents the joint prediction $\hat{\mathbf{p}}$ from falling into $\mathcal{U}$.

\paragraph{Composite Phonological Distance Metric}
Given a discrete predicted feature vector $\hat{\mathbf{p}} \in \mathcal{P}$ and a target handshape vector $\mathbf{p}^{(h)} \in \mathcal{P}_{\text{LSC}}$, we measure their dissimilarity using a composite distance function
$d(\hat{\mathbf{p}}, \mathbf{p}^{(h)}) = \sum_{c \in \mathcal{C}} d_c(\hat{p}_c, p_c^{(h)})$ 
where $\mathcal{C} = \{\text{sel}, \text{flex}, \text{sprd}, \text{th\_pos}, \text{th\_cnt}\}$ represents the set of feature categories.
The category distances $d_c$ are defined as follows.
First, $d_{\text{sel}}$ is the Hamming distance between the two selected-finger representations, where a mismatch (present vs. absent) for any of the four individual fingers ($i, m, r, p$) adds $1$ to the distance.
Second, $d_{\text{flex}}$ is defined over the four flexion values (\textit{FullyOpen}, \textit{Flat}, \textit{Curved/Bent}, and \textit{FullyClosed}), assigning a distance of $1$ to every distinct pair except the extreme transition $\text{FullyOpen} \leftrightarrow \text{FullyClosed}$, which costs $2$. 
Finally, $d_{\text{sprd}}$, $d_{\text{th\_pos}}$, and $d_{\text{th\_cnt}}$ assign a distance of $0$ for matching values\footnote{We include the \texttt{NA} class as a distinct, valid state for Spread and Thumb contact.} and $1$ for mismatches.

\paragraph{Candidate Selection and Evaluation}
The model outputs discrete categorical class predictions for each channel. Rather than requiring an exact vector match, we map the predicted feature vector $\hat{\mathbf{p}}$ to a set of candidate target handshapes $\hat{\mathcal{H}} \subset \mathcal{H}$ by retrieving all classes whose feature vectors minimize $d(\hat{\mathbf{p}}, \mathbf{p}^{(h)})$: $\hat{\mathcal{H}} = \operatorname*{argmin}_{h \in \mathcal{H}} d(\hat{\mathbf{p}}, \mathbf{p}^{(h)})$. Because multiple handshapes can tie for the minimum distance, $\hat{\mathcal{H}}$ is a candidate set containing one or more class labels. We evaluate performance using \textit{Expected Accuracy}. Let $h^* \in \mathcal{H}$ denote the true target class. Sample-level expected accuracy $A(h^*, \hat{\mathcal{H}})$ measures the probability of selecting $h^*$ under a uniform random choice from $\hat{\mathcal{H}}$, defined as $A(h^*, \hat{\mathcal{H}}) = 1/|\hat{\mathcal{H}}|$ if $h^* \in \hat{\mathcal{H}}$, and $0$ otherwise.
Table~\ref{tab:expected_acc_example} illustrates a prediction $\hat{\mathbf{p}} \in \mathcal{U}$ mapped to a candidate set $\hat{\mathcal{H}}$ of three equidistant target classes ($d=1$)\footnote{See \href{https://thesignhub.eu/grammar/lsc?tag=82}{SignHub's Finger Configuration}.}. This instance scores $1/3$ if $h^* \in \hat{\mathcal{H}}$, and $0$ otherwise.

\begin{table}[tb]
\centering
\small
\renewcommand{\arraystretch}{1.1}
\resizebox{\columnwidth}{!}{%
\begin{tabular}{@{} l l c @{}}
\toprule
\textbf{Candidates ($h$)} & \textbf{Phonological Features ($\mathbf{p}^{(h)}$)} & $\bm{d(\hat{\mathbf{p}}, \mathbf{p}^{(h)})}$ \\
\midrule
Prediction ($\hat{\mathbf{p}} \in \mathcal{U}$) & $\{i, m, r\}$, FullyOpen, $1$, Open, $0$ & -- \\
\midrule
\texttt{\href{https://ww1.thesignhub.eu/api/rest/retrievePublic?mediaUuid=UUID-MD-44689f2c-e65c-47e5-b212-aed5fa671424}{8-extended}} & {\color{red} $\{i, m\}$}, FullyOpen, $1$, Open, $0$ & $1$ \\
\texttt{\href{https://ww1.thesignhub.eu/api/rest/retrievePublic?mediaUuid=UUID-MD-f17d5081-da5f-4808-af2d-c09122634f65}{5-extended}} & {\color{red} $\{i, m, r, p\}$}, FullyOpen, $1$, Open, $0$ & $1$ \\
\texttt{\href{https://ww1.thesignhub.eu/api/rest/retrievePublic?mediaUuid=UUID-MD-a57e682c-6e1f-41ee-9a96-33d6307712ae}{3-extended}} & $\{i, m, r\}$, FullyOpen, $1$, {\color{red} Closed}, $0$ & $1$ \\
\midrule
\texttt{\href{https://ww1.thesignhub.eu/api/rest/retrievePublic?mediaUuid=UUID-MD-a37564ff-e1b5-4d87-a706-d9ae8f1558c0}{2-extended}} & {\color{red} $\{i, m\}$}, FullyOpen, $1$, {\color{red} Closed}, $0$ & $2$ \\
\texttt{\href{https://ww1.thesignhub.eu/api/rest/retrievePublic?mediaUuid=UUID-MD-dd71337f-d7e3-4171-beec-47938bcdc79f}{4-extended}} & {\color{red} $\{i, m, r, p\}$}, FullyOpen, $1$, {\color{red} Closed}, $0$ & $2$ \\
\texttt{\href{https://ww1.thesignhub.eu/api/rest/retrievePublic?mediaUuid=UUID-MD-d803100a-a9fa-4c3a-8737-7a87974c2880}{5-curved-open}} & {\color{red} $\{i, m, r, p\}$}, {\color{red} Curved/Bent}, $1$, Open, $0$ & $2$ \\
\texttt{\href{https://ww1.thesignhub.eu/api/rest/retrievePublic?mediaUuid=UUID-MD-86a5e2da-c309-46b8-a3f3-7ee322a652ff}{8-curved-open}} & {\color{red} $\{i, m\}$}, {\color{red} Curved/Bent}, $1$, Open, $0$ & $2$ \\
\texttt{\href{https://ww1.thesignhub.eu/api/rest/retrievePublic?mediaUuid=UUID-MD-ac620a15-c1f6-4503-b282-ba7e8a5a3b2e}{B-extended}} & {\color{red} $\{i, m, r, p\}$}, FullyOpen, {\color{red} $0$}, Open, $0$ & $2$ \\
\bottomrule
\end{tabular}%
}
\caption{Candidate selection for a prediction in $\mathcal{U}$. Three target classes tie at minimum distance $d=1$, forming $\hat{\mathcal{H}}$. Next-nearest classes ($d=2$) are excluded. Mismatched features are in red.}
\label{tab:expected_acc_example}
\end{table}

\subsection{Mitigating Recording-Format Disparities}
\label{subsec:mitigating_recording_format_disparities}
The LSC benchmark consists of uniform 16:9 1080p full-body studio footage recorded at 50 fps. In contrast, the training corpora rely on in-the-wild recordings---webcam video for Sem-Lex and smartphone footage for PopSign---with both training sets having a median frame rate near 30 fps.
To isolate cross-lingual errors from these recording-format disparities, we modify the LSC Benchmark clips to match the training capture geometries and subsample them to lower frame rates.
To address spatial disparities, we extract a static bounding box centered on the median shoulder position for each target LSC clip, scaling it to match the training corpus medians. Specifically, the target normalized shoulder width and vertical $y$-position are aligned with Sem-Lex's 4:3 landscape webcam format (denoted as webcam) and PopSign's 3:4 portrait smartphone format (denoted as phone).
Finally, to address temporal disparities, frame rate subsampling is applied to the LSC target clips after feature extraction using strides of 2 (25 fps) and 3 (16.7 fps). Integer strides avoid frame interpolation; 25\,fps approaches the training corpora's frame-rate medians, while 16.7\,fps probes robustness below them.
Note that harmonization uses no target-side supervision: the crop is computed independently for each clip from its own pose sequence together with source-corpus constants, without labels or any fitting across the test set, and is therefore applicable to any input video at inference time.

\section{Results}
\label{sec:results}

In this section, we empirically evaluate the cross-lingual transferability of phonological representations from ASL to LSC. We first verify our baseline source-task reproductions and address pose extraction artifacts. Next, we evaluate zero-shot transfer performance on the LSC benchmark for the three model architectures (MLP, SL-GCN, and SHuBERT) and the two training corpora (Sem-Lex and PopSign). Finally, we present a series of ablations to isolate the key drivers of transfer performance, examining the impact of recording-format alignment, auxiliary supervision heads, and SHuBERT's input channels and pretraining.

\paragraph{Source-Task Reproduction and Pose Correction}
\label{sec:source}

Table~\ref{tab:repro_results_summary} compares our reproduced performance on Sem-Lex against the metrics reported for SL-GCN \cite{kezar2023sem} and SHuBERT \cite{shubert}.
While our SHuBERT reproduction closely matches the published results, reaching 80.2\% test macro Recall@1 versus the reported 79.4\%, SL-GCN gloss recognition exhibits a noticeable gap. Training on the released Sem-Lex pose files yields 57.1\% Top-1 accuracy, underperforming the reported 66.6\% by 9.5 percentage points.
However, when the model is trained with the poses we extracted from the source videos—instead of Sem-Lex released poses—and following the identical recipe, gloss top-1 accuracy increases to 72.2\%, surpassing the published baseline rather than falling 9.5 points below it. Then, we fine-tune it for joint gloss and phonological feature prediction and improve the phonological feature average from 78.7\% to 87.1\%. The shortfall could be a property of the released data.
Elsewhere in this paper the SL-GCN Sem-Lex rows refer to the model trained with re-extracted poses.
Further reproduction details are provided in Appendix \ref{app:reproduction_results}.

\begin{table}[tb]
\centering
\setlength{\tabcolsep}{4pt}
\resizebox{\columnwidth}{!}{%

\begin{tabular}{lrrrrrr}
\toprule
 & \multicolumn{3}{c}{\textbf{Gloss Top-1 (\%)}~$\uparrow$} & \multicolumn{3}{c}{\textbf{16-Feat.\ Ph.\ (\%)}~$\uparrow$} \\
\cmidrule(lr){2-4} \cmidrule(lr){5-7}
\textbf{Model} & \textbf{Ref.} & \textbf{Ours} & \textbf{$\Delta$} & \textbf{Ref.} & \textbf{Ours} & \textbf{$\Delta$} \\
\midrule
SL-GCN (released poses) & 66.6 & 57.1 & -9.5 & 85.0 & 78.7 & -6.3 \\
SL-GCN (re-posed) & 66.6 & 72.2 & +5.6 & 85.0 & 87.1 & +2.1 \\
\midrule
SHuBERT & -- & -- & -- & 79.4 & 80.2 & +0.9 \\
\bottomrule
\end{tabular}%
}
\caption{Reproduction fidelity of published Sem-Lex-trained source models on the test set. Gloss Top-1: top-1 sign gloss classification accuracy; 16-Feat.\ Ph.: mean accuracy across all 16 phonological features. Ref. denotes published baselines from \citet{kezar2023sem} for SL-GCN and \citet{shubert} for SHuBERT. The $\Delta$ columns give absolute percentage point differences between our reproduction and the reference.}
\label{tab:repro_results_summary}
\end{table}

\paragraph{Zero-shot transfer to LSC}
We train all three model architectures on PopSign and Sem-Lex using the full set of 16 phonological features. This choice maintains consistency with prior work and we hypothesize that additional feature targets provide auxiliary supervision without degrading performance, as ablated later in this section. However, since our cross-lingual handshape recognition approach relies only on five handshape-related phonological features, we evaluate performance using the mean accuracy across these five phonological features (Ph.\%~$\uparrow$) and the expected handshape accuracy (Hs.\%~$\uparrow$). All models are evaluated across three random seeds both in-domain on the corresponding ASL test splits and zero-shot on LSC, reporting mean performance alongside standard deviations (Table~\ref{tab:indomain_vs_transfer}).

As anticipated, all models score significantly lower in the zero-shot LSC setting than in-domain, with both metrics exhibiting similar trends. Training on Sem-Lex transfers better to unseen LSC handshapes than training on PopSign, particularly for the MLP (11.8\% vs.\ 27.3\% Hs.) and SHuBERT (13.9\% vs.\ 18.3\% Hs.) models. For SL-GCN, the two training corpora yield indistinguishable zero-shot performance (8.3\% vs.\ 8.6\% Hs.).
Notably, while the MLP is the weakest model in-domain,
it emerges as the \emph{strongest} zero-shot model on LSC, leading both transfer metrics (62.6\% Ph., 27.3\% Hs.). We hypothesize that high-capacity spatio-temporal models overfit to source-dataset biases and pose extraction artifacts.

\begin{table}[tb]
\centering
\setlength{\tabcolsep}{2pt}
\resizebox{\columnwidth}{!}{%
\begin{tabular}{llcccc}
\toprule
 & & \multicolumn{2}{c}{\textbf{ASL test}} & \multicolumn{2}{c}{\textbf{LSC zero-shot}} \\
\cmidrule(lr){3-4} \cmidrule(lr){5-6}
\textbf{Train} & \textbf{Model} & \textbf{Ph.\,(\%)}~$\uparrow$ & \textbf{Hs.\,(\%)}~$\uparrow$ & \textbf{Ph.\,(\%)}~$\uparrow$ & \textbf{Hs.\,(\%)}~$\uparrow$ \\
\midrule
\multirow{3}{*}{\textbf{PopSign}} & MLP & 76.5$\pm$0.1 & 55.4$\pm$0.1 & 53.6$\pm$1.6 & 11.8$\pm$1.7 \\
 & SL-GCN & \textbf{98.2$\pm$0.1} & \textbf{96.5$\pm$0.1} & 52.3$\pm$0.7 & 8.3$\pm$2.4 \\
 & SHuBERT & 94.3$\pm$0.1 & 89.7$\pm$0.3 & \textbf{56.7$\pm$1.5} & \textbf{13.9$\pm$0.7} \\
\midrule
\multirow{3}{*}{\textbf{Sem-Lex}} & MLP & 75.2$\pm$0.2 & 48.8$\pm$0.2 & \textbf{62.6$\pm$1.7} & \textbf{27.3$\pm$1.2} \\
 & SL-GCN & \textbf{90.0$\pm$0.1} & \textbf{78.4$\pm$0.3} & 47.9$\pm$2.6 & 8.6$\pm$3.7 \\
 & SHuBERT & 81.7$\pm$0.2 & 64.9$\pm$0.2 & 59.9$\pm$1.3 & 18.3$\pm$4.2 \\
\bottomrule
\end{tabular}
}
\caption{In-domain ASL vs.\ zero-shot LSC transfer performance. Three model families are trained on PopSign and Sem-Lex, then evaluated on their respective ASL test splits and zero-shot on the LSC benchmark. Ph.: mean accuracy across handshape phonological features; Hs.: expected handshape accuracy. Cells report the mean $\pm$ standard deviation across random seeds. \textbf{Bold} indicates the best-performing model within each training corpus block.}
\label{tab:indomain_vs_transfer}
\end{table}

\paragraph{Mitigating the recording-format gap}
As outlined in Section~\ref{subsec:mitigating_recording_format_disparities}, we re-evaluated each model under recording conditions closer to its training distribution to isolate cross-lingual transfer from recording-format disparities. To maintain a strict zero-shot source-only setup, harmonization is determined by the training corpus alone: each model is evaluated with spatial cropping matching its capture geometry and the frame rate closest to its training median---webcam at 25\,fps for Sem-Lex-trained models and phone at 25\,fps for PopSign-trained ones (shaded cells in Table~\ref{tab:crop_fps_ablation_mean}). Under this corpus-matched setting, the Sem-Lex MLP reaches 80.0\% Ph.\ / 54.5\% Hs., the Sem-Lex SHuBERT 65.6\% / 22.1\%, and the PopSign SL-GCN 70.4\% / 29.5\%---gains of $+17.4$ / $+27.2$, $+5.7$ / $+3.8$, and $+18.1$ / $+21.2$ points (Ph.\ / Hs.), respectively, over their uncropped 50\,fps baselines (Table~\ref{tab:indomain_vs_transfer}).

For completeness and as an ablation of both factors, Table~\ref{tab:crop_fps_ablation_mean} also reports performance across the full grid of crop and frame-rate combinations. Spatial cropping drives the vast majority of the gains: compared to uncropped baselines at matched frame rates, webcam and phone cropping yield average improvements of $+11.1$ / $+11.3$ and $+10.5$ / $+10.6$ (Ph.\ / Hs.), respectively, whereas frame-rate reduction offers marginal benefits ($\le 0.7$ points averaged across models; see Table~\ref{tab:crop_trends} in Appendix~\ref{app:mitigating_recording_format}).

Crucially, tuning the configuration on the target benchmark---rather than adhering to a strict zero-shot protocol---yields minimal gain. The grid optimum coincides with the corpus-matched setting for the strongest model (Sem-Lex MLP). For the remaining models, the grid optima exceed the corpus-matched configuration by at most 2.4 Ph.\ / 2.7 Hs.\ points (e.g., Sem-Lex SHuBERT at webcam, 50\,fps: 66.1\% / 23.5\%; PopSign SL-GCN at phone, 16.7\,fps: 72.5\% / 31.5\%). Thus, the observed improvements hold under a strictly zero-shot, source-only setup.

Appendix~\ref{app:per_feature_analysis} decomposes these results per phonological feature: the corpus-matched harmonization gains concentrate on selected fingers and spread, while flexion remains the weakest feature.

\begin{table}[t]
\centering
\setlength{\tabcolsep}{3pt}
\resizebox{\columnwidth}{!}{%
\begin{tabular}{lllccc|ccc}
\toprule
 & & & \multicolumn{3}{c}{\textbf{Ph.\,(\%)}~$\uparrow$} & \multicolumn{3}{c}{\textbf{Hs.\,(\%)}~$\uparrow$} \\
\cmidrule(lr){4-6} \cmidrule(lr){7-9}
 & \textbf{Model} & \textbf{Crop} & \textbf{50} & \textbf{25} & \textbf{16.7} & \textbf{50} & \textbf{25} & \textbf{16.7} \\
\midrule
\multirow{9}{*}{\rotatebox[origin=c]{90}{\textbf{PopSign}}} & \multirow{3}{*}{MLP} & no crop & 53.6 & 50.4 & 51.9 & 11.8 & 10.4 & 12.0 \\
 &  & webcam & 71.9 & \textbf{72.6} & 71.9 & 25.3 & 26.9 & \textbf{27.8} \\
 &  & phone & 70.7 & \cellcolor{gray!15}70.2 & 69.5 & 26.0 & \cellcolor{gray!15}26.4 & 25.9 \\
\cmidrule(l){2-9}
 & \multirow{3}{*}{SL-GCN} & no crop & 52.3 & 55.0 & 54.7 & 8.3 & 11.8 & 9.9 \\
 &  & webcam & 67.0 & 68.2 & 69.3 & 19.8 & 22.4 & 20.7 \\
 &  & phone & 70.4 & \cellcolor{gray!15}70.4 & \textbf{72.5} & 26.0 & \cellcolor{gray!15}29.5 & \textbf{31.5} \\
\cmidrule(l){2-9}
 & \multirow{3}{*}{SHuBERT} & no crop & 56.7 & 55.6 & 54.8 & 13.9 & 13.4 & 14.5 \\
 &  & webcam & 55.6 & 56.2 & 56.8 & 14.6 & 15.8 & 15.1 \\
 &  & phone & 55.1 & \cellcolor{gray!15}\textbf{56.9} & 56.4 & 14.9 & \cellcolor{gray!15}\textbf{15.8} & 13.5 \\
\midrule
\multirow{9}{*}{\rotatebox[origin=c]{90}{\textbf{Sem-Lex}}} & \multirow{3}{*}{MLP} & no crop & 62.6 & 60.5 & 62.9 & 27.3 & 22.4 & 25.0 \\
 &  & webcam & 79.7 & \cellcolor{gray!15}\textbf{80.0} & 79.8 & 53.8 & \cellcolor{gray!15}\textbf{54.5} & 52.9 \\
 &  & phone & 76.5 & 76.9 & 76.3 & 42.7 & 43.7 & 42.8 \\
\cmidrule(l){2-9}
 & \multirow{3}{*}{SL-GCN} & no crop & 47.9 & 50.2 & 51.6 & 8.6 & 9.6 & 10.3 \\
 &  & webcam & 56.3 & \cellcolor{gray!15}57.2 & 57.8 & 13.1 & \cellcolor{gray!15}14.5 & 15.9 \\
 &  & phone & 54.3 & 58.2 & \textbf{59.1} & 13.1 & \textbf{17.2} & 15.2 \\
\cmidrule(l){2-9}
 & \multirow{3}{*}{SHuBERT} & no crop & 59.9 & 58.5 & 58.2 & 18.3 & 15.7 & 16.3 \\
 &  & webcam & \textbf{66.1} & \cellcolor{gray!15}65.6 & 65.0 & \textbf{23.5} & \cellcolor{gray!15}22.1 & 23.5 \\
 &  & phone & 64.6 & 63.9 & 63.6 & 22.7 & 20.2 & 22.5 \\
\bottomrule
\end{tabular}%
}
\caption{Zero-shot evaluation on the LSC benchmark of video-crop (none, 4:3 webcam, 3:4 phone) and frame-rate (50, 25, 16.7) ablations for models trained on PopSign and Sem-Lex. Ph.: mean accuracy of five handshape phonological features; Hs.: expected handshape accuracy. \textbf{Bold} indicates the best Ph.\ and Hs.\ result per model configuration. \colorbox{gray!15}{Shaded} cells mark the corpus-matched harmonization configuration---crop matching the training corpus, 25\,fps (nearest to the training median)---fixed without consulting LSC results.}
\label{tab:crop_fps_ablation_mean}
\end{table}

\paragraph{Target phonological heads}
\label{par:target_phonological_heads}
To ablate the effect of auxiliary supervision, we retrained each model family on both ASL corpora to predict only the five handshape phonological features alongside the handshape class, rather than all 16 phonological heads. Holding model architectures and random seeds fixed, we present these comparative results in Table~\ref{tab:heads_ablation} (Appendix \ref{app:ablation_target_heads}).
Removing auxiliary supervision yields no systematic benefits. In-domain, the shallow MLP is largely unaffected ($\le 0.3$ percentage point shift), while deeper architectures suffer minor-to-moderate declines in handshape accuracy, particularly on Sem-Lex (SHuBERT $-4.8$, SL-GCN $-3.1$). On zero-shot LSC transfer, performance fluctuates directionlessly (e.g., SHuBERT Ph.\ $+2.9$ vs.\ SL-GCN Ph.\ $-4.7$ on Sem-Lex) with overlapping seed standard deviations in nearly all cases. We conclude that while auxiliary heads offer mild regularization for deep models in-domain, they neither systematically aid nor degrade cross-lingual transfer.

\paragraph{SHuBERT ablations}
We ablate SHuBERT along two axes (input channels and pretraining), varying one factor at a time against the all-channel pretrained reference (Table~\ref{tab:shubert_ablation}).
\textbf{Input channels:} SHuBERT fuses four positional streams: face, left hand, right hand, and body posture. We evaluate configurations retaining all four channels, the two hand streams only, and the body stream only. Dropped channels are masked using SHuBERT's pretrained mask embedding. 
Restricting the input to hand channels matches the all-channel reference with performance gaps in-domain falling within the standard deviation, demonstrating that face and body pose contribute minimally to handshape recognition.
Conversely, the body-only model collapses: without hand signals, it predicts the training majority class, yielding a constant-predictor floor (marked $\dagger\dagger$ in Table~\ref{tab:shubert_ablation}).
\textbf{Pretraining:} training SHuBERT from scratch without self-supervised pretraining confirms that SSL pretraining provides substantial performance gains across all setups. On Sem-Lex, pretraining boosts in-domain phonological feature macro accuracy by $+7.4$ points and handshape accuracy by $+14.2$ points, whereas on PopSign the corresponding gains are $+3.7$ and $+6.6$ points. Consequently, zero-shot transfer to LSC exhibits a corresponding degradation when removing pre-training.

\begin{table}[tb]
\centering
\setlength{\tabcolsep}{4pt}
\resizebox{\columnwidth}{!}{%
\begin{tabular}{llcccc}
\toprule
 & & \multicolumn{2}{c}{\textbf{ASL test}} & \multicolumn{2}{c}{\textbf{LSC zero-shot}} \\
\cmidrule(lr){3-4} \cmidrule(lr){5-6}
 & \textbf{Ablation} & \textbf{Ph.\,(\%)}~$\uparrow$ & \textbf{Hs.\,(\%)}~$\uparrow$ & \textbf{Ph.\,(\%)}~$\uparrow$ & \textbf{Hs.\,(\%)}~$\uparrow$ \\
\midrule
\multirow{4}{*}{\rotatebox[origin=c]{90}{\textbf{PopSign}}} & all channels (reference) & \textbf{94.3$\pm$0.1} & \textbf{89.7$\pm$0.3} & \textbf{56.7$\pm$1.5} & \textbf{13.9$\pm$0.7} \\
 & hands-only & \textbf{94.3$\pm$0.1} & 89.4$\pm$0.3 & 56.2$\pm$2.0 & \textbf{13.9$\pm$3.8} \\
 & body-only & 55.4$\pm$0.0 & 12.3$\pm$0.0 & 50.3$\pm$0.0$^{\dagger\dagger}$ & 2.7$\pm$0.0$^{\dagger\dagger}$ \\
 & no pre-training & 90.6$\pm$0.5 & 83.1$\pm$0.9 & 50.6$\pm$4.6 & 8.6$\pm$3.5 \\
\midrule
\multirow{4}{*}{\rotatebox[origin=c]{90}{\textbf{Sem-Lex}}} & all channels (reference) & 81.7$\pm$0.2 & \textbf{64.9$\pm$0.2} & 59.9$\pm$1.3 & 18.3$\pm$4.2 \\
 & hands-only & \textbf{81.8$\pm$0.1} & 64.2$\pm$0.6 & \textbf{64.4$\pm$3.2} & \textbf{23.5$\pm$5.8} \\
 & body-only & 55.0$\pm$0.0 & 11.0$\pm$1.1 & 50.3$\pm$0.0$^{\dagger\dagger}$ & 2.7$\pm$0.0$^{\dagger\dagger}$ \\
 & no pre-training & 74.3$\pm$0.3 & 50.7$\pm$1.0 & 58.4$\pm$1.4 & 14.9$\pm$1.5 \\
\bottomrule
\end{tabular}
}
\caption{Input channel and pre-training ablation for SHuBERT fine-tuned on PopSign and Sem-Lex, evaluated in-domain (ASL) and zero-shot (LSC). Ph.: mean phonological feature accuracy; Hs.: expected handshape accuracy. Cells report mean $\pm$ std over seeds; \textbf{bold} denotes best result per training corpus. $^{\dagger\dagger}$Degenerate runs emitting constant majority-class predictions.}
\label{tab:shubert_ablation}
\end{table}

\section{Discussion}
\label{sec:discussion}
\paragraph{Cross-Lingual Transfer and Phonological Universality}
Our findings show that zero-shot cross-lingual transfer from ASL to LSC is viable once domain gaps in recording format such as spatial cropping are explicitly harmonized. This supports the hypothesis that phonological features capture foundational articulatory properties that transcend individual sign languages. However, important scope limitations remain. Our evaluation is restricted to a single historically related language pair ($\text{ASL} \rightarrow \text{LSC}$) and a benchmark containing two videos per handshape from a single signer across 37 handshape classes (out of more than 43 described in LSC, excluding fingerspelling, numerals, and classifiers).
In particular, signer-independent generalization within LSC cannot be established from this benchmark, and robustness to uncontrolled target recording conditions remains untested.
Although our decoding framework remains mathematically agnostic to the target set $\mathcal{P}_{\text{LSC}}$, future work will evaluate this approach across larger multi-signer datasets and typologically unrelated sign language families to test cross-lingual universality.

\paragraph{Model Capacity, Dataset Biases, and Practical Efficiency}
A key insight from our experiments is the counterintuitive strength of the shallow MLP, which outperforms deeper architectures in zero-shot transfer. We hypothesize that high-capacity spatio-temporal models (such as SHuBERT and SL-GCN) overfit to source-dataset biases and pose-extraction artifacts. 
Because the MLP lacks temporal mechanics and receives only dominant-hand features, its capacity bottleneck prevents it from learning non-transferable contextual cues, yielding superior zero-shot generalization. Crucially, our ablations show that SHuBERT can be stripped to hand-only input streams and supervised on only six handshape-relevant heads without sacrificing accuracy. This stream pruning significantly reduces the annotation burden and computational overhead required to deploy phonological recognizers for handshape prediction in new target languages.

\paragraph{Static Phonological Representation}
\label{par:phonological_representation_limits_and_extensions}
Although the 5-dimensional phonological feature vector successfully grounds categorical decoding, its formulation reveals structural limits as observed by \citet{sehyr2021asl}. First, the feature mapping is non-injective: the five phonological categories used in this work fail to capture unselected finger flexion, causing distinct handshapes to map to identical phonological feature vectors. Second, reducing signs to a single static handshape simplifies dynamic signing reality, overlooking handshape transitions over time as well as non-dominant hand configurations. Extending the shared feature space to parameterize unselected fingers, dynamic temporal trajectories, and two-handed signs to the feature space would expand cross-lingual transfer across the complete manual phonological spectrum.

\paragraph{Expected Accuracy and Decoding Ties}
These representational constraints are directly manifested during evaluation, defining the expected handshape accuracy as a diagnostic metric rather than a standard classification performance score. Because candidate ties are resolved uniformly at random, even an ideal feature predictor achieves a theoretical ceiling of 94.6\% (reflecting 35 unique canonical feature configurations across 37 target classes). Such ties stem from two distinct mechanisms: \emph{structural ties} between handshape class pairs sharing identical phonological feature vectors, and \emph{distance ties}, where a prediction lies equidistant from multiple valid LSC handshape classes. Future work can resolve distance ties by leveraging the model's per-feature confidences---ensuring a single, deterministic prediction---and disambiguate structural ties only through supplementary information, such as unselected finger flexion features.

\section{Conclusions}
\label{sec:conclusions}

In this work, we introduced the first zero-shot cross-lingual framework for sign language handshape recognition, bridging American Sign Language (ASL) and Catalan Sign Language (LSC) through a shared, linguistically grounded phonological feature space. By decomposing handshapes into five phonological features and mapping predictions via a composite distance metric, our method enables direct target-language classification without requiring target-language phonological training labels. 

Evaluated across three distinct model architectures (MLP, SL-GCN, and SHuBERT) and two source corpora (PopSign and Sem-Lex), our experiments on a single-signer LSC benchmark show that zero-shot transfer is viable—achieving up to 80.0\% phonological feature accuracy and 54.5\% expected handshape accuracy. Crucially, we show that isolating true cross-lingual transfer requires first harmonizing recording-format disparities such as spatial cropping.
Our experiments reveal that low-capacity models (MLP) offer superior zero-shot generalization, providing evidence that model complexity does not necessarily translate into better cross-lingual transfer; we hypothesize that high-capacity models overfit source-dataset spatio-temporal biases. Complex multimodal transformers (SHuBERT), meanwhile, can be pruned to hand-only streams without loss of handshape performance. 
These results support the hypothesis that phonological decomposition can provide a language-agnostic representation bridge: for the language pair studied, this work substantially reduces target-language annotation requirements and opens new avenues for scaling sign language technologies to underserved sign communities and low-resource sign languages worldwide.

\section*{Acknowledgments}
This work is part of Project PID2025-173709NB-I00 (MuReLA), funded by MICIU/AEI/10.13039/501100011033. It was also supported by the Maria de Maeztu Units of Excellence Programme (CEX2021-001195-M, funded by MICIU/AEI/10.13039/501100011033), the PULSAR project (Ref. PID2025-173459NB-C22, financed by MICIU/AEI/10.13039/501100011033 and FEDER/UE), and VISTASL (2024-1-EL01-KA220-HED-000257847), funded by the EU's Erasmus+ programme.
M.~G. was supported by the Joan Oró predoctoral programme (FI) of the Department of Research and Universities of the Generalitat de Catalunya (file no.\ 2024~FI-1~01038), co-funded by the European Social Fund Plus.
G.~H. acknowledges support from the Serra Húnter Programme
(Generalitat de Catalunya) as a Serra Húnter Associate Professor.
We acknowledge the EuroHPC Joint Undertaking for awarding us access to MareNostrum5 at BSC (Spain) and Leonardo at CINECA (Italy).

We would like to thank Josep Blat and Laia Tarrés for their advice and support, Lee Kezar for early discussions on the topic, UPF's LSC-Lab and Lali Ribera for their willingness to collaborate, and in particular Alexandra Navarrete for guiding our understanding of the phonology of Catalan Sign Language.

\bibliography{custom}

@misc{WFD_stats,
    author = {{World Federation of the Deaf}},
    title = {FAQ: Deaf Communities and Sign Languages},
    year = {2026},
    url = {https://wfdeaf.org/contact/faqs/},
    note = {Accessed: 2026-08-21}
}

@inproceedings{bragg2019sign,
  title={Sign language recognition, generation, and translation: An interdisciplinary perspective},
  author={Bragg, Danielle and Koller, Oscar and Bellard, Mary and Berke, Larwan and Boudreault, Patrick and Braffort, Annelies and Caselli, Naomi and Huenerfauth, Matt and Kacorri, Hernisa and Verhoef, Tessa and others},
  booktitle={Proceedings of the 21st international ACM SIGACCESS conference on computers and accessibility},
  pages={16--31},
  year={2019}
}

@misc{iec2025corpuslsc,
  author       = {{Institut d'Estudis Catalans}},
  title        = {Corpus de refer{\`e}ncia de la llengua de signes catalana ({LSC}) ({CORPUS LSC})},
  year         = {2025},
  howpublished = {\url{https://corpuslsc.iec.cat/}},
  url          = {https://corpuslsc.iec.cat/},
  doi          = {10.2436/10.2500.22.1}
}

@inproceedings{camgoz2018neural,
  title={Neural sign language translation},
  author={Camgoz, Necati Cihan and Hadfield, Simon and Koller, Oscar and Ney, Hermann and Bowden, Richard},
  booktitle={2018 IEEE/CVF Conference on Computer Vision and Pattern Recognition},
  pages={7784--7793},
  year={2018},
  organization={Ieee}
}

@inproceedings{camgoz2020sign,
  title={Sign language transformers: Joint end-to-end sign language recognition and translation},
  author={Camgoz, Necati Cihan and Koller, Oscar and Hadfield, Simon and Bowden, Richard},
  booktitle={Proceedings of the IEEE/CVF conference on computer vision and pattern recognition},
  pages={10023--10033},
  year={2020}
}

@inproceedings{forster2012rwth,
  title={RWTH-PHOENIX-weather: A large vocabulary sign language recognition and translation corpus.},
  author={Forster, Jens and Schmidt, Christoph and Hoyoux, Thomas and Koller, Oscar and Zelle, Uwe and Piater, Justus H and Ney, Hermann},
  booktitle={LREC},
  volume={9},
  pages={3785--3789},
  year={2012}
}

@inproceedings{forster2014extensions,
  title={Extensions of the Sign Language Recognition and Translation Corpus RWTH-PHOENIX-Weather.},
  author={Forster, Jens and Schmidt, Christoph and Koller, Oscar and Bellgardt, Martin and Ney, Hermann},
  booktitle={LREC},
  pages={1911--1916},
  year={2014}
}

@inproceedings{zhou2023gloss,
  title={Gloss-free sign language translation: Improving from visual-language pretraining},
  author={Zhou, Benjia and Chen, Zhigang and Clap{\'e}s, Albert and Wan, Jun and Liang, Yanyan and Escalera, Sergio and Lei, Zhen and Zhang, Du},
  booktitle={2023 IEEE/CVF International Conference on Computer Vision (ICCV)},
  pages={20814--20824},
  year={2023},
  organization={IEEE}
}

@inproceedings{lin2023gloss,
  title={Gloss-free end-to-end sign language translation},
  author={Lin, Kezhou and Wang, Xiaohan and Zhu, Linchao and Sun, Ke and Zhang, Bang and Yang, Yi},
  booktitle={Proceedings of the 61st Annual Meeting of the Association for Computational Linguistics (Volume 1: Long Papers)},
  pages={12904--12916},
  year={2023}
}

@inproceedings{muller2023considerations,
  title={Considerations for meaningful sign language machine translation based on glosses},
  author={M{\"u}ller, Mathias and Jiang, Zifan and Moryossef, Amit and Gonzales, Annette Rios and Ebling, Sarah},
  booktitle={Proceedings of the 61st Annual Meeting of the Association for Computational Linguistics (Volume 2: Short Papers)},
  pages={682--693},
  year={2023}
}

@article{stokoe1960sign,
      title={Sign Language Structure: An Outline of the Visual Communication Systems of the {A}merican Deaf},
      author={Stokoe, William C.},
      journal={Studies in Linguistics, Occasional Papers},
      volume={8},
      year={1960},
}

@book{brentari1998prosodic,
      title={A Prosodic Model of Sign Language Phonology},
      author={Brentari, Diane},
      year={1998},
      publisher={MIT Press},
      address={Cambridge, MA},
}

@book{battison1978lexical,
      title={Lexical Borrowing in American Sign Language},
      author={Battison, Robbin},
      year={1978},
      publisher={Linstok Press},
      address={Silver Spring, MD},
}

@incollection{navarrete2020phonology,
  author    = {Navarrete-Gonz{\'a}lez, Alexandra},
  title     = {Phonology: 1. Sublexical Structure},
  booktitle = {A Grammar of Catalan Sign Language ({LSC})},
  editor    = {Quer, Josep and Barber{\`a}, Gemma},
  series    = {{SIGN-HUB} Sign Language Grammar Series},
  edition   = {1st},
  year      = {2020},
  url       = {https://thesignhub.eu/grammar/lsc?tag=71},
  urldate   = {2021-10-31}
}

@book{quer2020grammar,
  editor    = {Quer, Josep and Barber{\`a}, Gemma},
  title     = {A Grammar of Catalan Sign Language ({LSC})},
  series    = {{SIGN-HUB} Sign Language Grammar Series},
  edition   = {1st},
  year      = {2020},
  url       = {http://www.thesignhub.eu/grammar/lsc},
  urldate   = {2021-10-31}
}

@article{wittmann1991classification,
  title={Classification linguistique des langues sign{\'e}es non vocalement},
  author={Wittmann, Henri},
  journal={Revue qu{\'e}b{\'e}coise de linguistique th{\'e}orique et appliqu{\'e}e},
  volume={10},
  number={1},
  pages={215--288},
  year={1991},
  publisher={Association qu{\'e}b{\'e}coise de linguistique}
}

@article{uthus2023youtube,
  title={Youtube-asl: A large-scale, open-domain american sign language-english parallel corpus},
  author={Uthus, Dave and Tanzer, Garrett and Georg, Manfred},
  journal={Advances in Neural Information Processing Systems},
  volume={36},
  pages={29029--29047},
  year={2023}
}

@inproceedings{tanzer2025youtube,
  title={Youtube-sl-25: A large-scale, open-domain multilingual sign language parallel corpus},
  author={Tanzer, Garrett and Zhang, Biao},
  booktitle={International Conference on Learning Representations},
  volume={2025},
  pages={81921--81934},
  year={2025}
}

@inproceedings{koller2016deep,
  title={Deep hand: How to train a cnn on 1 million hand images when your data is continuous and weakly labelled},
  author={Koller, Oscar and Ney, Hermann and Bowden, Richard},
  booktitle={Proceedings of the IEEE conference on computer vision and pattern recognition},
  pages={3793--3802},
  year={2016}
}

@article{caselli2017asl,
  title={{ASL-LEX}: A lexical database of American Sign Language},
  author={Caselli, Naomi K and Sehyr, Zed Sevcikova and Cohen-Goldberg, Ariel M and Emmorey, Karen},
  journal={Behavior research methods},
  volume={49},
  number={2},
  pages={784--801},
  year={2017},
  publisher={Springer}
}

@article{sehyr2021asl,
  title={The {ASL-LEX} 2.0 Project: A database of lexical and phonological properties for 2,723 signs in American Sign Language},
  author={Sehyr, Zed Sevcikova and Caselli, Naomi and Cohen-Goldberg, Ariel M and Emmorey, Karen},
  journal={The Journal of Deaf Studies and Deaf Education},
  volume={26},
  number={2},
  pages={263--277},
  year={2021},
  publisher={Oxford University Press}
}

@inproceedings{kezar2023sem,
  title={The sem-lex benchmark: Modeling asl signs and their phonemes},
  author={Kezar, Lee and Thomason, Jesse and Caselli, Naomi and Sehyr, Zed and Pontecorvo, Elana},
  booktitle={Proceedings of the 25th International ACM SIGACCESS Conference on Computers and Accessibility},
  pages={1--10},
  year={2023}
}

@misc{hochgesang2019signbank,
  title = {{ASL Signbank}},
  author = {Hochgesang, Julie A. and Crasborn, Onno and Lillo-Martin, Diane},
  year = {2019},
  howpublished = {New Haven, CT: Haskins Lab, Yale University},
  url = {https://aslsignbank.haskins.yale.edu/}
}

@article{starner2023popsign,
  title={Popsign asl v1. 0: An isolated american sign language dataset collected via smartphones},
  author={Starner, Thad and Forbes, Sean and So, Matthew and Martin, David and Sridhar, Rohit and Deshpande, Gururaj and Sepah, Sam and Shahryar, Sahir and Bhardwaj, Khushi and Kwok, Tyler and others},
  journal={Advances in Neural Information Processing Systems},
  volume={36},
  pages={184--196},
  year={2023}
}

@misc{chow2023google,
  title     = {{Google - Isolated Sign Language Recognition}},
  author    = {Chow, Ashley and Cameron, Glenn and Sherwood, Mark and Culliton, Phil and Sepah, Sam and Dane, Sohier and Starner, Thad},
  year      = {2023},
  url       = {https://kaggle.com/competitions/asl-signs},
  publisher = {Kaggle}
}

@inproceedings{jiang2021skeleton,
  title={Skeleton aware multi-modal sign language recognition},
  author={Jiang, Songyao and Sun, Bin and Wang, Lichen and Bai, Yue and Li, Kunpeng and Fu, Yun},
  booktitle={2021 IEEE/CVF Conference on Computer Vision and Pattern Recognition Workshops (CVPRW)},
  pages={3408--3418},
  year={2021},
  organization={IEEE}
}

@article{lugaresi2019mediapipe,
  title={Mediapipe: A framework for building perception pipelines},
  author={Lugaresi, Camillo and Tang, Jiuqiang and Nash, Hadon and McClanahan, Chris and Uboweja, Esha and Hays, Michael and Zhang, Fan and Chang, Chuo-Ling and Yong, Ming Guang and Lee, Juhyun and others},
  journal={arXiv preprint arXiv:1906.08172},
  year={2019}
}

@inproceedings{tornay2020towards,
  title={Towards multilingual sign language recognition},
  author={Tornay, Sandrine and Razavi, Marzieh and Doss, Mathew Magimai-},
  booktitle={ICASSP 2020-2020 IEEE International Conference on Acoustics, Speech and Signal Processing (ICASSP)},
  pages={6309--6313},
  year={2020},
  organization={IEEE}
}

@inproceedings{inoue-etal-2026-resource,
    title = "A Resource and Evaluation Method for Phonological Continuity in {J}apanese {S}ign {L}anguage",
    author = "Inoue, Jundai and Hara, Daisuke and Miwa, Makoto",
    booktitle = "Proceedings of the Fifteenth Language Resources and Evaluation Conference",
    month = may, year = "2026",
    address = "Palma de Mallorca, Spain",
    publisher = "ELRA Language Resource Association",
    pages = "9514--9524",
    doi = "10.63317/4p22nojyxbxa"
}

@article{bilge2024crosslingual,
  title = {Cross-lingual few-shot sign language recognition},
  author = {Bilge, Yunus Can and Ikizler-Cinbis, Nazli and Cinbis, Ramazan Gokberk},
  journal = {Pattern Recognition},
  volume = {151},
  pages = {110374},
  year = {2024},
  publisher = {Elsevier},
  doi = {10.1016/j.patcog.2024.110374}
}

@inproceedings{Handshape-GNN,
    title = "Improving Handshape Representations for Sign Language Processing: A Graph Neural Network Approach",
    author = "Carbo, Alessa  and
      Nalisnick, Eric",
    editor = "Christodoulopoulos, Christos  and
      Chakraborty, Tanmoy  and
      Rose, Carolyn  and
      Peng, Violet",
    booktitle = "Proceedings of the 2025 Conference on Empirical Methods in Natural Language Processing",
    month = nov,
    year = "2025",
    address = "Suzhou, China",
    publisher = "Association for Computational Linguistics",
    url = "https://aclanthology.org/2025.emnlp-main.1483/",
    doi = "10.18653/v1/2025.emnlp-main.1483",
    pages = "29122--29135",
    ISBN = "979-8-89176-332-6"
}

@misc{shubert,
      title={{SHuBERT}: Self-Supervised Sign Language Representation Learning via Multi-Stream Cluster Prediction}, 
      author={Shester Gueuwou and Xiaodan Du and Greg Shakhnarovich and Karen Livescu and Alexander H. Liu},
      year={2025},
      eprint={2411.16765},
      archivePrefix={arXiv},
      primaryClass={cs.CL},
      url={https://arxiv.org/abs/2411.16765}, 
}

\clearpage
\appendix
\onecolumn
\newpage
\section{LSC Handshape Inventory and Feature Mapping}
\label{sec:appendix-handshapes}
LSC handshapes were derived from the \href{https://thesignhub.eu/grammar/lsc?tag=82}{SignHub LSC Grammar: Finger Configuration}. We inferred selected fingers from the finger configuration table, relying either on explicit mentions in the handshape names or on standard number and letter conventions. Configuration values correspond to the original header row, while spread was manually annotated following ASL-LEX rules.

Table~\ref{tab:handshape-inventory} presents the complete inventory of 43 LSC handshapes alongside their original SignHub annotations and mapped ASL-LEX features. The deterministic alignment procedure used to link these two feature spaces is detailed in Section~\ref{subsec:mapping_lsc_handshapes_to_asl-lex}.
Note that spread distinguishes class \emph{5} (spread) from \emph{B} (unspread) exclusively for \emph{extended} and \emph{curved open} configurations. The remaining configurations collapse into a single \emph{5B} class. We encode this merged class as unspread based on SignHub's LSC finger configuration illustrations and the benchmark clips.

As noted by \citet{sehyr2021asl} and discussed in Section~\ref{par:phonological_representation_limits_and_extensions}, the five ASL-LEX handshape features---\textit{selected fingers}, \textit{flexion}, \textit{spread}, \textit{thumb position}, and \textit{thumb contact}---cannot distinguish all handshapes, as certain classes differ exclusively in the flexion of unselected fingers.

\begin{table*}[h!]
\centering
\footnotesize
\resizebox{0.99\textwidth}{!}{%
\setlength{\tabcolsep}{4pt}
\begin{tabular}{ccccccccc}
\toprule
 & \multicolumn{3}{c}{\textbf{SignHub-inferred LSC annotations}} & \multicolumn{5}{c}{\textbf{ASL-LEX 2.0 phonological features}} \\
\cmidrule(lr){2-4}\cmidrule(lr){5-9}
\textbf{LSC Handshape} & \textbf{Sel.\ fingers} & \textbf{Configuration} & \textbf{Spread} & \textbf{Sel.\ fingers} & \textbf{Flexion} & \textbf{Spread} & \textbf{Thumb pos.} & \textbf{Thumb cont.} \\
\midrule
6-extended & t & extended & NA & t & FullyOpen & NA & Open & NA \\
6-curved\_open & t & curved open & NA & t & Curved/Bent & NA & Open & NA \\
6-closed & t & closed & NA & t & FullyClosed & NA & Open & NA \\
\addlinespace[2pt]
1-extended & i & extended & NA & i & FullyOpen & NA & Closed & 0 \\
1-curved\_open & i & curved open & NA & i & Curved/Bent & NA & Closed & 0 \\
1-closed & i & closed & NA & i & FullyClosed & NA & Closed & 1 \\
\addlinespace[2pt]
middle-extended & m & extended & NA & m & FullyOpen & NA & Closed & 0 \\
middle-flat\_open & m & flat open & NA & m & Flat & NA & Closed & 0 \\
\addlinespace[2pt]
i-extended & p & extended & NA & p & FullyOpen & NA & Closed & 0 \\
i-curved\_open$^{*}$ & p & curved open & NA & p & Curved/Bent & NA & Closed & 0 \\
\addlinespace[2pt]
7-extended & ti & extended & NA & i & FullyOpen & NA & Open & 0 \\
7-flat\_open$^{\dagger}$ & ti & flat open & NA & i & Flat & NA & Open & 0 \\
7-flat\_closed$^{\ddagger}$ & ti & flat closed & NA & i & Flat & NA & Open & 1 \\
7-curved\_open & ti & curved open & NA & i & Curved/Bent & NA & Open & 0 \\
\addlinespace[2pt]
index+thumb-flat\_open$^{\dagger}$ & ti & flat open & NA & i & Flat & NA & Open & 0 \\
index+thumb-flat\_closed$^{\ddagger}$ & ti & flat closed & NA & i & Flat & NA & Open & 1 \\
index+thumb-curved\_closed & ti & curved closed & NA & i & Curved/Bent & NA & Open & 1 \\
\addlinespace[2pt]
middle+thumb-flat\_open$^{*}$ & tm & flat open & NA & m & Flat & NA & Open & 0 \\
middle+thumb-flat\_closed$^{*}$ & tm & flat closed & NA & m & Flat & NA & Open & 1 \\
middle+thumb-curved\_closed$^{*}$ & tm & curved closed & NA & m & Curved/Bent & NA & Open & 1 \\
\addlinespace[2pt]
2-extended & im & extended & spread & im & FullyOpen & 1 & Closed & 0 \\
2-curved\_open & im & curved open & spread & im & Curved/Bent & 1 & Closed & 0 \\
2-curved\_closed & im & curved closed & spread & im & Curved/Bent & 1 & Closed & 1 \\
\addlinespace[2pt]
n-extended & im & extended & unspread & im & FullyOpen & 0 & Closed & 0 \\
n-curved\_open$^{*}$ & im & curved open & unspread & im & Curved/Bent & 0 & Closed & 0 \\
\addlinespace[2pt]
Y-extended & tp & extended & NA & p & FullyOpen & NA & Open & 0 \\
\addlinespace[2pt]
u-extended & ip & extended & NA & ip & FullyOpen & NA & Closed & 0 \\
\addlinespace[2pt]
8-extended & tim & extended & spread & im & FullyOpen & 1 & Open & 0 \\
8-flat\_closed & tim & flat closed & spread & im & Flat & 0 & Open & 1 \\
8-curved\_open & tim & curved open & spread & im & Curved/Bent & 1 & Open & 0 \\
\addlinespace[2pt]
4-extended & imrp & extended & spread & imrp & FullyOpen & 1 & Closed & 0 \\
4-flat\_open & imrp & flat open & spread & imrp & Flat & 0 & Closed & 0 \\
\addlinespace[2pt]
5-extended & timrp & extended & spread & imrp & FullyOpen & 1 & Open & 0 \\
5-curved\_open & timrp & curved open & spread & imrp & Curved/Bent & 1 & Open & 0 \\
\addlinespace[2pt]
b-extended & timrp & extended & unspread & imrp & FullyOpen & 0 & Open & 0 \\
b-curved\_open & timrp & curved open & unspread & imrp & Curved/Bent & 0 & Open & 0 \\
\addlinespace[2pt]
5B-flat\_open & timrp & flat open & unspread$^{**}$ & imrp & Flat & 0 & Open & 0 \\
5B-flat\_closed & timrp & flat closed & unspread$^{**}$ & imrp & Flat & 0 & Open & 1 \\
5B-curved\_closed & timrp & curved closed & unspread$^{**}$ & imrp & Curved/Bent & 0 & Open & 1 \\
5B-closed & timrp & closed & unspread$^{**}$ & imrp & FullyClosed & NA & Open & 1 \\
\addlinespace[2pt]
pinkie+middle+thumb-extended & tmp & extended & NA & mp & FullyOpen & NA & Open & 0 \\
\addlinespace[2pt]
3-extended & imr & extended & spread & imr & FullyOpen & 1 & Closed & 0 \\
3-curved\_open$^{*}$ & imr & curved open & spread & imr & Curved/Bent & 1 & Closed & 0 \\
\bottomrule
\end{tabular}%
}
\caption{Mapping of the 43 LSC handshape classes \cite{navarrete2020phonology} to ASL-LEX 2.0 phonological features, alongside their SignHub-inferred annotations. Asterisks ($^{*}$) mark the six handshape classes lacking visual clips in the benchmark. The double asterisk ($^{**}$) denotes collapsed \emph{5B} classes encoded as unspread (see text). Classes sharing a dagger ($^{\dagger}$, $^{\ddagger}$) possess identical ASL-LEX feature configurations, differing only in the flexion of unselected fingers.}
\label{tab:handshape-inventory}
\end{table*}

\clearpage
\section{Phonological Feature Inventories Across Datasets}
\label{app:phonological_feature_inventories_across_datasets}
To analyze the structural granularity of the feature spaces across datasets, Table~\ref{tab:feature_class_counts} contrasts the class counts across each phonological feature for Sem-Lex, PopSign, and the LSC Benchmark. Table~\ref{tab:feature_values} provides an itemized comparison of the value inventories for the five primary handshape features, explicitly denoting merged classes (e.g., \textit{Curved/Bent}) and structural gaps between ASL-LEX and LSC.

Comparing these label spaces highlights a key structural divergence in \textit{Selected Fingers}: the combination \textit{mp} (middle and pinkie) is attested in LSC (e.g., in the \textit{pinkie+middle+thumb-extended} handshape) but absent from the ASL-LEX training set. Consequently, our model cannot directly predict the exact \textit{mp} value during zero-shot evaluation. Nevertheless, because target handshape retrieval relies on a distance metric over the combined feature space, \textit{pinkie+middle+thumb-extended} can still be successfully recovered within the candidate set. To systematically handle unobserved finger combinations in future work, finger selection could be decomposed into independent per-digit binary features rather than treating joint configurations as discrete atomic classes.

\begin{table}[h!]
\centering
\resizebox{0.55\textwidth}{!}{%
\begin{tabular}{lrrr}
\toprule
\textbf{Phonological Feature} & \textbf{Sem-Lex} & \textbf{PopSign} & \textbf{LSC Benchmark} \\
\midrule
\multicolumn{4}{l}{\emph{Other phonological features}} \\
Major Location & 5 & 5 & -- \\
Minor Location & 37 & 25 & -- \\
Second Minor Location & 37 & 16 & -- \\
Contact & 2 & 2 & -- \\
Sign Type & 6 & 5 & -- \\
Repeated Movement & 2 & 2 & -- \\
Path Movement & 8 & 7 & -- \\
Wrist Twist & 2 & 2 & -- \\
Spread Change & 3 & 3 & -- \\
Nondominant Handshape & 56 & 25 & -- \\
\midrule
\multicolumn{4}{l}{\emph{Handshape phonological features}} \\
Selected Fingers & 12 & 9 & 9 \\
Flexion & 8 & 7 & 4 \\
Spread & 3 & 3 & 3 \\
Thumb Position & 2 & 2 & 2 \\
Thumb Contact & 3 & 3 & 3 \\
\midrule
\multicolumn{4}{l}{\emph{Handshape class}} \\
Handshape & 58 & 37 & 37 \\
\bottomrule
\end{tabular}
}
\caption{Number of classes per phonological feature in each dataset's label space. Sem-Lex uses the full ASL-LEX 2.0 coding; PopSign's space is reduced to the values attested among its 250 glosses; the LSC Benchmark column is the canonical target space of the five handshape features in ASL-LEX format (Curved and Bent flexion merged) plus the 37 handshape classes -- the remaining features have no LSC annotation (--). Counts include the NA class where the annotation has structurally-absent values (Table~\ref{tab:feature_values} lists the value inventories of the five handshape features).}
\label{tab:feature_class_counts}
\end{table}

\begin{table}[h!]
\centering
\small
\begin{tabular}{lp{6.2cm}p{5.2cm}}
\toprule
\textbf{Feature} & \textbf{ASL-LEX 2.0 values} & \textbf{LSC values (ASL-LEX format)} \\
\midrule
Selected Fingers & t, i, m, \textbf{r}, p, im, ip, \textbf{mr}, imr, \textbf{imp}, \textbf{mrp}, imrp & t, i, m, p, im, ip, \textbf{mp}, imr, imrp \\
\addlinespace[2pt]
Flexion & FullyOpen, Flat, \textbf{Curved}, \textbf{Bent}, FullyClosed, \textbf{Stacked}, \textbf{Crossed}, \textbf{NA} & FullyOpen, Flat, \textbf{Curved/Bent}, FullyClosed \\
\addlinespace[2pt]
Spread & 0, 1, NA & 0, 1, NA \\
\addlinespace[2pt]
Thumb Position & Open, Closed & Open, Closed \\
\addlinespace[2pt]
Thumb Contact & 0, 1, NA & 0, 1, NA \\
\bottomrule
\end{tabular}
\caption{Value inventories of the five phonological features strictly related to handshape: the ASL-LEX 2.0 space the models are trained on vs the reduced LSC target space in ASL-LEX format. \textbf{Bold} marks the changes: ASL-LEX values with no identical LSC counterpart (Crossed, Stacked, and flexion NA have no LSC realisation; Curved and Bent collapse into the merged Curved/Bent class; four selected-finger combinations are unattested in LSC) and LSC values absent from the ASL-LEX space (the merged Curved/Bent; mp, attested in LSC but not in ASL-LEX). NA is a real class, marking structurally-absent values, never missing supervision.}
\label{tab:feature_values}
\end{table}

\clearpage

\twocolumn
\section{Reproduction Results}
\label{app:reproduction_results}
To ensure the reliability of our baseline implementations, we reproduce experiments on both SL-GCN \cite{kezar2023sem} and SHuBERT \cite{shubert}. 
Table~\ref{tab:slgcn_table5_repro} reports the reproduction fidelity for SL-GCN across two distinct pose inputs: the official pose files shipped with the Sem-Lex release versus pose sequences extracted directly from the raw dataset videos using our pipeline. Evaluating on the official released poses reveals a systematic performance drop (averaging $-6.3$ percentage points), whereas our re-extracted pose sequences reliably match and slightly exceed original paper metrics ($+2.1$ percentage points on average across phonological tasks).

\begin{table}[tb]
\centering
\resizebox{\columnwidth}{!}{%
\begin{tabular}{l c cc cc}
\toprule
 & \multicolumn{5}{c}{\textbf{Test Top-1 Accuracy (\%)}~$\uparrow$} \\
\cmidrule(lr){2-6}
 & & \multicolumn{2}{c}{\textbf{Released Poses}} & \multicolumn{2}{c}{\textbf{Re-posed}} \\
\cmidrule(lr){3-4} \cmidrule(lr){5-6}
\textbf{Feature} & \textbf{Ref.} & \textbf{Ours} & \textbf{$\Delta$} & \textbf{Ours} & \textbf{$\Delta$} \\
\midrule
\multicolumn{6}{l}{\emph{Gloss ISR (Tab.~4/6): the feature heads' encoder}} \\
Gloss top-1 & 66.6 & 57.1 & -9.5 & 72.2 & +5.6 \\
Gloss top-3 & 81.5 & 71.1 & -10.4 & 85.6 & +4.1 \\
\midrule
\multicolumn{6}{l}{\emph{Phonological features (Tab.~5, multitask)}} \\
Major Location & 87.5 & 79.4 & -8.1 & 88.4 & +0.9 \\
Minor Location & 78.1 & 67.3 & -10.8 & 80.0 & +1.9 \\
2nd Minor Location & 77.2 & 69.2 & -8.0 & 80.4 & +3.2 \\
Contact & 88.6 & 85.1 & -3.5 & 90.7 & +2.1 \\
Sign Type & 87.9 & 80.8 & -7.1 & 88.8 & +0.9 \\
Repeated Movement & 85.4 & 81.5 & -3.9 & 88.6 & +3.2 \\
Path Movement & 75.4 & 69.7 & -5.7 & 80.4 & +5.0 \\
Wrist Twist & 92.6 & 89.5 & -3.1 & 93.5 & +0.9 \\
Spread Change & 89.5 & 84.3 & -5.2 & 91.2 & +1.7 \\
Nondominant Handshape & 81.7 & 72.4 & -9.3 & 82.9 & +1.2 \\
Selected Fingers & 90.2 & 84.6 & -5.6 & 92.3 & +2.1 \\
Flexion & 81.0 & 73.8 & -7.2 & 83.3 & +2.3 \\
Spread & 88.0 & 81.7 & -6.3 & 89.6 & +1.6 \\
Thumb Position & 91.5 & 87.7 & -3.8 & 92.6 & +1.1 \\
Thumb Contact & 91.1 & 86.4 & -4.7 & 92.2 & +1.1 \\
Handshape & 74.7 & 65.6 & -9.1 & 78.4 & +3.7 \\
\textbf{Average} & \textbf{85.0} & \textbf{78.7} & \textbf{-6.3} & \textbf{87.1} & \textbf{+2.1} \\
\bottomrule
\end{tabular}
}
\caption{SL-GCN reproduction fidelity on the test split. We compare the results reported by Sem-Lex \cite{kezar2023sem} against our reproduction runs. Our models are evaluated on two pose sets: the original \emph{Released Poses} provided by the benchmark, and \emph{Re-extracted}, which denotes the poses we independently extracted from the same videos. The $\Delta$ columns indicate the absolute percentage point difference between our reproduction and the baseline.}
\label{tab:slgcn_table5_repro}
\end{table}

Table~\ref{tab:shubert_table9_repro} evaluates our fine-tuned SHuBERT baseline against the published Sem-Lex test results. Our reproduced SHuBERT model consistently matches or marginally outperforms published per-feature Recall@1 scores in all 16 phonological dimensions, achieving an average performance gain of $+0.86$ percentage points.
\begin{table}[tb]
\centering
\setlength{\tabcolsep}{1.5pt}
\resizebox{0.75\columnwidth}{!}{%
\begin{tabular}{l c c c}
\toprule
 & \multicolumn{3}{c}{\textbf{Test Recall@1 (\%)}~$\uparrow$} \\
\cmidrule(lr){2-4}
\textbf{Phonological Feature} & \textbf{Ref.} & \textbf{Ours} & \textbf{$\Delta$} \\
\midrule
Major Location & 84.77 & 85.03 & +0.26 \\
Minor Location & 71.30 & 71.73 & +0.43 \\
2nd Minor Location & 73.28 & 73.64 & +0.36 \\
Contact & 86.84 & 87.35 & +0.51 \\
Sign Type & 84.64 & 84.66 & +0.02 \\
Repeated Movement & 82.65 & 84.01 & +1.36 \\
Path Movement & 72.75 & 72.92 & +0.17 \\
Wrist Twist & 90.58 & 91.48 & +0.90 \\
Spread Change & 81.60 & 82.85 & +1.25 \\
Nondominant Handshape & 76.32 & 76.72 & +0.40 \\
Selected Fingers & 79.53 & 81.04 & +1.51 \\
Flexion & 72.64 & 74.38 & +1.74 \\
Spread & 79.42 & 80.38 & +0.96 \\
Thumb Position & 86.04 & 87.03 & +0.99 \\
Thumb Contact & 84.74 & 85.76 & +1.02 \\
Handshape & 62.93 & 64.89 & +1.96 \\
\midrule
\textbf{Average} & \textbf{79.38} & \textbf{80.24} & \textbf{+0.86} \\
\bottomrule
\end{tabular}
}
\caption{SHuBERT reproduction fidelity evaluated on the Sem-Lex test set. We compare the per-feature Recall@1 results reported in the original SHuBERT baseline \cite{shubert} against our fine-tuned reproduction. The $\Delta$ column indicates the absolute percentage point difference between our reproduction and the baseline.}
\label{tab:shubert_table9_repro}
\end{table}

\section{Further results for mitigation of recording-format gap}
\label{app:mitigating_recording_format}
To provide a full accounting of stability and individual factor contributions in our domain-adaptation experiments, this section presents detailed breakdowns of the cropping format and frame-rate ablations.
Table~\ref{tab:crop_fps_ablation_std} extends Table~\ref{tab:crop_fps_ablation_mean} from the main text by reporting sample standard deviations across three random seeds for every cell in the $3 \times 3$ grid of crop configurations (uncropped, 4:3 webcam, and 3:4 phone portrait) and frame rates (50\,fps, 25\,fps, and 16.7\,fps). Table~\ref{tab:crop_trends} isolates the marginal treatment effects ($\Delta$) of spatial cropping relative to uncropped inputs, as well as temporal frame-rate reductions relative to native 50\,fps video.

\textbf{Spatial alignment drives gains.} Spatial cropping yields substantial performance improvements for all models ($+10.5\%$--$+11.3\%$ average gain), peaking at $+28.9\%$ for MLP handshape accuracy on Sem-Lex. This shows pose-based models are sensitive to video framing, though SHuBERT is less affected by uncropped inputs. Conversely, frame-rate reductions (from 50\,fps to 25 or 16.7\,fps) have a negligible impact ($\le +0.7\%$ average change). This confirms that spatial domain shift—rather than temporal resolution—is the primary bottleneck in zero-shot cross-dataset transfer.

\begin{table*}[p]
\centering
\setlength{\tabcolsep}{3pt}
\resizebox{0.9\textwidth}{!}{%
\begin{tabular}{lllccc|ccc}
\toprule
 & & & \multicolumn{3}{c}{\textbf{Ph.\,(\%)}~$\uparrow$} & \multicolumn{3}{c}{\textbf{Hs.\,(\%)}~$\uparrow$} \\
\cmidrule(lr){4-6} \cmidrule(lr){7-9}
 & \textbf{Model} & \textbf{Crop} & \textbf{50\,fps} & \textbf{25\,fps} & \textbf{16.7\,fps} & \textbf{50\,fps} & \textbf{25\,fps} & \textbf{16.7\,fps} \\
\midrule
\multirow{9}{*}{\rotatebox[origin=c]{90}{\textbf{PopSign}}} & \multirow{3}{*}{MLP} & no crop & 53.6$\pm$1.6 & 50.4$\pm$2.6 & 51.9$\pm$2.1 & 11.8$\pm$1.7 & 10.4$\pm$2.8 & 12.0$\pm$2.9 \\
 &  & webcam crop & 71.9$\pm$2.1 & \textbf{72.6$\pm$2.0} & 71.9$\pm$2.1 & 25.3$\pm$2.0 & 26.9$\pm$2.8 & \textbf{27.8$\pm$3.7} \\
 &  & phone crop & 70.7$\pm$2.3 & \cellcolor{gray!15}70.2$\pm$1.1 & 69.5$\pm$1.3 & 26.0$\pm$1.9 & \cellcolor{gray!15}26.4$\pm$1.8 & 25.9$\pm$1.4 \\
\cmidrule(l){2-9}
 & \multirow{3}{*}{SL-GCN} & no crop & 52.3$\pm$0.7 & 55.0$\pm$1.0 & 54.7$\pm$2.3 & 8.3$\pm$2.4 & 11.8$\pm$0.5 & 9.9$\pm$1.7 \\
 &  & webcam crop & 67.0$\pm$1.4 & 68.2$\pm$1.8 & 69.3$\pm$0.7 & 19.8$\pm$2.2 & 22.4$\pm$3.3 & 20.7$\pm$2.2 \\
 &  & phone crop & 70.4$\pm$1.1 & \cellcolor{gray!15}70.4$\pm$0.9 & \textbf{72.5$\pm$0.4} & 26.0$\pm$1.6 & \cellcolor{gray!15}29.5$\pm$2.7 & \textbf{31.5$\pm$1.0} \\
\cmidrule(l){2-9}
 & \multirow{3}{*}{SHuBERT} & no crop & 56.7$\pm$1.5 & 55.6$\pm$1.0 & 54.8$\pm$2.3 & 13.9$\pm$0.7 & 13.4$\pm$1.1 & 14.5$\pm$1.6 \\
 &  & webcam crop & 55.6$\pm$1.5 & 56.2$\pm$1.5 & 56.8$\pm$1.1 & 14.6$\pm$1.4 & 15.8$\pm$1.0 & 15.1$\pm$0.7 \\
 &  & phone crop & 55.1$\pm$1.4 & \cellcolor{gray!15}\textbf{56.9$\pm$1.3} & 56.4$\pm$1.2 & 14.9$\pm$0.2 & \cellcolor{gray!15}\textbf{15.8$\pm$1.6} & 13.5$\pm$1.8 \\
\midrule
\multirow{9}{*}{\rotatebox[origin=c]{90}{\textbf{Sem-Lex}}} & \multirow{3}{*}{MLP} & no crop & 62.6$\pm$1.7 & 60.5$\pm$1.9 & 62.9$\pm$1.6 & 27.3$\pm$1.2 & 22.4$\pm$0.8 & 25.0$\pm$1.2 \\
 &  & webcam crop & 79.7$\pm$1.2 & \cellcolor{gray!15}\textbf{80.0$\pm$0.9} & 79.8$\pm$0.7 & 53.8$\pm$2.6 & \cellcolor{gray!15}\textbf{54.5$\pm$3.6} & 52.9$\pm$3.3 \\
 &  & phone crop & 76.5$\pm$1.2 & 76.9$\pm$1.1 & 76.3$\pm$0.4 & 42.7$\pm$1.9 & 43.7$\pm$2.2 & 42.8$\pm$2.3 \\
\cmidrule(l){2-9}
 & \multirow{3}{*}{SL-GCN} & no crop & 47.9$\pm$2.6 & 50.2$\pm$3.1 & 51.6$\pm$4.4 & 8.6$\pm$3.7 & 9.6$\pm$2.1 & 10.3$\pm$4.0 \\
 &  & webcam crop & 56.3$\pm$4.3 & \cellcolor{gray!15}57.2$\pm$3.3 & 57.8$\pm$0.9 & 13.1$\pm$4.8 & \cellcolor{gray!15}14.5$\pm$1.2 & 15.9$\pm$0.5 \\
 &  & phone crop & 54.3$\pm$5.3 & 58.2$\pm$5.3 & \textbf{59.1$\pm$1.0} & 13.1$\pm$3.6 & \textbf{17.2$\pm$4.4} & 15.2$\pm$1.7 \\
\cmidrule(l){2-9}
 & \multirow{3}{*}{SHuBERT} & no crop & 59.9$\pm$1.3 & 58.5$\pm$0.7 & 58.2$\pm$1.8 & 18.3$\pm$4.2 & 15.7$\pm$3.2 & 16.3$\pm$4.6 \\
 &  & webcam crop & \textbf{66.1$\pm$2.3} & \cellcolor{gray!15}65.6$\pm$2.0 & 65.0$\pm$1.7 & \textbf{23.5$\pm$5.3} & \cellcolor{gray!15}22.1$\pm$1.8 & 23.5$\pm$5.1 \\
 &  & phone crop & 64.6$\pm$0.7 & 63.9$\pm$1.3 & 63.6$\pm$1.9 & 22.7$\pm$2.5 & 20.2$\pm$1.1 & 22.5$\pm$3.6 \\
\bottomrule
\end{tabular}
}
\caption{Video-crop $\times$ frame-rate ablation on the LSC benchmark with seed spreads: the appendix sibling of Table~\ref{tab:crop_fps_ablation_mean}. Models are trained on PopSign and Sem-Lex and evaluated zero-shot on the LSC Benchmark. Ph. is the mean accuracy over the five handshape phonological features; Hs. is the expected handshape accuracy; decoding is constrained to the LSC inventory. The \emph{Crop} column gives the type of image cropping applied to the LSC clips (none, webcam-style 4:3, or phone-style 3:4 portrait), with each model's input re-extracted from the cropped video; the numeric sub-columns give the video frame rate in fps (native 50, subsampled 25 and 16.7). \textbf{Bold} marks, for each trained model, its best Ph.\ and best Hs.\ cell over the whole grid; \colorbox{gray!15}{shaded} cells mark the corpus-matched harmonization configuration (crop matching the training corpus, 25\,fps), fixed without consulting LSC results. Cells are mean\,$\pm$\,sample sd over three seeds.}
\label{tab:crop_fps_ablation_std}
\end{table*}

\begin{table*}[p]
\centering
\setlength{\tabcolsep}{2pt}
\resizebox{0.9\textwidth}{!}{%
\begin{tabular}{llcccccccc}
\toprule
 & & \multicolumn{4}{c}{\textbf{Phonological feature macro $\Delta$ (\%)}~$\uparrow$} & \multicolumn{4}{c}{\textbf{Expected handshape $\Delta$ (\%)}~$\uparrow$} \\
\cmidrule(lr){3-6} \cmidrule(lr){7-10}
 & \textbf{Model} & \textbf{webcam} & \textbf{phone} & \textbf{25} & \textbf{16.7} & \textbf{webcam} & \textbf{phone} & \textbf{25} & \textbf{16.7} \\
\midrule
\multirow{4}{*}{\rotatebox[origin=c]{90}{\textbf{PopSign}}} & MLP & \textbf{+20.2$\pm$2.3} & +18.2$\pm$2.0 & -1.0$\pm$0.4 & -1.0$\pm$0.3 & \textbf{+15.3$\pm$3.6} & +14.7$\pm$3.1 & +0.2$\pm$0.3 & +0.9$\pm$0.9 \\
 & SL-GCN & +14.2$\pm$1.9 & \textbf{+17.1$\pm$1.2} & +1.3$\pm$1.3 & +2.3$\pm$1.0 & +11.0$\pm$2.8 & \textbf{+19.0$\pm$2.5} & +3.2$\pm$2.9 & +2.7$\pm$1.9 \\
 & SHuBERT & \textbf{+0.5$\pm$0.8} & \textbf{+0.5$\pm$0.5} & \textbf{+0.5$\pm$0.3} & +0.2$\pm$0.4 & \textbf{+1.2$\pm$1.7} & +0.8$\pm$1.0 & +0.5$\pm$0.4 & -0.1$\pm$1.1 \\
 & \textit{mean} & +11.6$\pm$10.1 & \textbf{+11.9$\pm$9.9} & +0.2$\pm$1.2 & +0.5$\pm$1.7 & +9.1$\pm$7.2 & \textbf{+11.5$\pm$9.5} & +1.3$\pm$1.7 & +1.2$\pm$1.4 \\
\midrule
\multirow{4}{*}{\rotatebox[origin=c]{90}{\textbf{Sem-Lex}}} & MLP & \textbf{+17.8$\pm$1.2} & +14.6$\pm$1.1 & -0.5$\pm$0.4 & +0.1$\pm$0.4 & \textbf{+28.9$\pm$2.4} & +18.2$\pm$1.4 & -1.0$\pm$0.5 & -1.0$\pm$0.7 \\
 & SL-GCN & +7.2$\pm$3.4 & \textbf{+7.3$\pm$4.7} & +2.3$\pm$1.9 & +3.3$\pm$2.1 & +5.0$\pm$3.1 & \textbf{+5.6$\pm$4.1} & +2.2$\pm$1.0 & +2.2$\pm$1.6 \\
 & SHuBERT & \textbf{+6.7$\pm$2.2} & +5.2$\pm$2.3 & -0.9$\pm$1.1 & -1.3$\pm$1.0 & \textbf{+6.3$\pm$0.8} & +5.1$\pm$2.3 & -2.2$\pm$1.7 & -0.7$\pm$2.3 \\
 & \textit{mean} & \textbf{+10.6$\pm$6.3} & +9.0$\pm$4.9 & +0.3$\pm$1.8 & +0.7$\pm$2.4 & \textbf{+13.4$\pm$13.4} & +9.6$\pm$7.4 & -0.3$\pm$2.3 & +0.1$\pm$1.8 \\
\midrule
\multirow{4}{*}{\rotatebox[origin=c]{90}{\textbf{Mean}}} & MLP & \textbf{+19.0$\pm$1.7} & +16.4$\pm$2.5 & -0.7$\pm$0.4 & -0.5$\pm$0.7 & \textbf{+22.1$\pm$9.6} & +16.4$\pm$2.5 & -0.4$\pm$0.9 & -0.1$\pm$1.3 \\
 & SL-GCN & +10.7$\pm$4.9 & \textbf{+12.2$\pm$6.9} & +1.8$\pm$0.7 & +2.8$\pm$0.7 & +8.0$\pm$4.2 & \textbf{+12.3$\pm$9.4} & +2.7$\pm$0.7 & +2.4$\pm$0.3 \\
 & SHuBERT & \textbf{+3.6$\pm$4.4} & +2.8$\pm$3.3 & -0.2$\pm$1.0 & -0.5$\pm$1.0 & \textbf{+3.8$\pm$3.6} & +2.9$\pm$3.0 & -0.8$\pm$1.9 & -0.4$\pm$0.5 \\
 & \textit{mean} & \textbf{+11.1$\pm$7.5} & +10.5$\pm$7.2 & +0.3$\pm$1.3 & +0.6$\pm$1.8 & \textbf{+11.3$\pm$9.9} & +10.6$\pm$7.7 & +0.5$\pm$2.0 & +0.7$\pm$1.5 \\
\bottomrule
\end{tabular}
}
\caption{Treatment effects of the crop $\times$ frame-rate grid, one row per trained model. \emph{webcam} and \emph{phone} are the crop's effect against the uncropped clips, averaged over the three frame rates; \emph{25} and \emph{16.7} are the frame rate's effect against native 50\,fps, averaged over the three crops. \emph{Mean} rows average the unrounded deltas of the rows in their scope (a corpus, a model family across both corpora, or all six models); their $\pm$ is the sd across the averaged rows, not a seed sd -- every other $\pm$ is a sample sd over three seeds.}
\label{tab:crop_trends}
\end{table*}

\clearpage
\twocolumn

\section{Per-Feature Error Analysis}
\label{app:per_feature_analysis}
Table~\ref{tab:per_feature_lsc} decomposes the zero-shot LSC results into the five phonological features, reporting the accuracy of each feature together with their mean (Ph.). Each model is evaluated zero-shot on the raw recordings (no crop, 50\,fps) and with its corpus-matched harmonization; the rows thus decompose exactly the aggregate values of Table~\ref{tab:crop_fps_ablation_mean}.
The \emph{majority class} row reports the accuracy of a constant predictor answering the benchmark's most frequent value per feature. Because our models are trained on ASL corpora whose class priors differ from the benchmark's and never observe its distribution, this row is a reference line rather than an expected floor.
Three patterns emerge. First, flexion---a four-way distinction of finger curvature---is the weakest or tied-weakest feature in every corpus-matched configuration. Second, harmonization gains concentrate on selected fingers and spread (e.g., $+24.3$ points each for the Sem-Lex MLP), while thumb position is nearly unaffected. Third, measured against the majority references, the learned signal is largest for selected fingers (88.3\% vs.\ a 27.0\% reference for the Sem-Lex MLP) and smallest for the two thumb features, whose accuracies stay within a few points of their references.
\begin{table}[t]
\centering
\setlength{\tabcolsep}{3pt}
\resizebox{\columnwidth}{!}{%
\begin{tabular}{lllcccccc}
\toprule
 & & & \multicolumn{5}{c}{\textbf{Per-feature accuracy (\%)}~$\uparrow$} & \\
\cmidrule(lr){4-8}
 & \textbf{Model} & \textbf{Config} & \textbf{SF} & \textbf{Flex.} & \textbf{Sprd.} & \textbf{Th.Pos.} & \textbf{Th.Cnt.} & \textbf{Ph.} \\
\midrule
\multicolumn{3}{l}{\emph{Majority class}} & 27.0 & 40.5 & 54.1 & 62.2 & 67.6 & 50.3 \\
\midrule
\multirow{6}{*}{\rotatebox[origin=c]{90}{\textbf{PopSign}}} & \multirow{2}{*}{MLP} & no crop, 50 & 47.7 & 44.6 & 64.9 & \textbf{64.4} & 46.4 & 53.6 \\
 &  & phone, 25 & \textbf{76.6} & \textbf{60.8} & \textbf{87.8} & 60.8 & \textbf{64.9} & \textbf{70.2} \\
\cmidrule(l){2-9}
 & \multirow{2}{*}{SL-GCN} & no crop, 50 & 49.5 & 40.1 & 53.6 & 64.9 & 53.2 & 52.3 \\
 &  & phone, 25 & \textbf{74.3} & \textbf{53.6} & \textbf{77.5} & \textbf{66.7} & \textbf{79.7} & \textbf{70.4} \\
\cmidrule(l){2-9}
 & \multirow{2}{*}{SHuBERT} & no crop, 50 & 39.6 & \textbf{44.1} & \textbf{64.9} & \textbf{67.6} & \textbf{67.1} & 56.7 \\
 &  & phone, 25 & \textbf{44.1} & \textbf{44.1} & 63.5 & 67.6 & 65.3 & \textbf{56.9} \\
\midrule
\multirow{6}{*}{\rotatebox[origin=c]{90}{\textbf{Sem-Lex}}} & \multirow{2}{*}{MLP} & no crop, 50 & 64.0 & 48.6 & 63.5 & 74.8 & 62.2 & 62.6 \\
 &  & webcam, 25 & \textbf{88.3} & \textbf{69.8} & \textbf{87.8} & \textbf{77.5} & \textbf{76.6} & \textbf{80.0} \\
\cmidrule(l){2-9}
 & \multirow{2}{*}{SL-GCN} & no crop, 50 & 43.7 & 33.8 & 45.0 & 61.3 & 55.9 & 47.9 \\
 &  & webcam, 25 & \textbf{57.7} & \textbf{41.9} & \textbf{58.6} & \textbf{63.1} & \textbf{64.9} & \textbf{57.2} \\
\cmidrule(l){2-9}
 & \multirow{2}{*}{SHuBERT} & no crop, 50 & 49.1 & 47.7 & 64.9 & 69.4 & 68.5 & 59.9 \\
 &  & webcam, 25 & \textbf{57.2} & \textbf{49.5} & \textbf{68.0} & \textbf{75.7} & \textbf{77.5} & \textbf{65.6} \\
\bottomrule
\end{tabular}%
}
\caption{Per-feature error analysis on the LSC benchmark: accuracy of each of the five phonological features and their mean (Ph.), computed over the 74 benchmark clips and three seeds. Each model is evaluated zero-shot on the raw recordings (no crop, native 50\,fps) and with corpus-matched harmonization (crop matching the training corpus, 25\,fps; shaded in Table~\ref{tab:crop_fps_ablation_mean}); the two settings differ only in label-free video pre-processing. Majority class: a constant predictor answering the benchmark's most frequent value per feature. Bold: best config per model and feature. SF: selected fingers.}
\label{tab:per_feature_lsc}
\end{table}

\begin{table}[tb]
\centering
\setlength{\tabcolsep}{4pt}
\resizebox{\columnwidth}{!}{%
\begin{tabular}{lllcccc}
\toprule
 & & & \multicolumn{2}{c}{\textbf{ASL Test}} & \multicolumn{2}{c}{\textbf{LSC Zero-Shot}} \\
\cmidrule(lr){4-5} \cmidrule(lr){6-7}
 & \textbf{Model} & \textbf{Setting} & \textbf{Ph.\,(\%)}~$\uparrow$ & \textbf{Hs.\,(\%)}~$\uparrow$ & \textbf{Ph.\,(\%)}~$\uparrow$ & \textbf{Hs.\,(\%)}~$\uparrow$ \\
\midrule
\multirow{6}{*}{\rotatebox[origin=c]{90}{\textbf{PopSign}}} & \multirow{2}{*}{MLP} & all16 & 76.5$\pm$0.1 & 55.4$\pm$0.1 & 53.6$\pm$1.6 & 11.8$\pm$1.7 \\
 &  & heads6 & \textbf{76.7$\pm$0.1} & \textbf{55.5$\pm$0.2} & \textbf{55.1$\pm$0.7} & \textbf{12.8$\pm$0.5} \\
\cmidrule(l){2-7}
 & \multirow{2}{*}{SL-GCN} & all16 & \textbf{98.2$\pm$0.1} & \textbf{96.5$\pm$0.1} & 52.3$\pm$0.7 & \textbf{8.3$\pm$2.4} \\
 &  & heads6 & 98.0$\pm$0.1 & 96.1$\pm$0.1 & \textbf{53.9$\pm$1.1} & 5.4$\pm$2.3 \\
\cmidrule(l){2-7}
 & \multirow{2}{*}{SHuBERT} & all16 & \textbf{94.3$\pm$0.1} & \textbf{89.7$\pm$0.3} & \textbf{56.7$\pm$1.5} & 13.9$\pm$0.7 \\
 &  & heads6 & 93.7$\pm$0.2 & 88.4$\pm$0.5 & 55.5$\pm$3.6 & \textbf{16.5$\pm$2.9} \\
\midrule
\multirow{6}{*}{\rotatebox[origin=c]{90}{\textbf{Sem-Lex}}} & \multirow{2}{*}{MLP} & all16 & 75.2$\pm$0.2 & \textbf{48.8$\pm$0.2} & \textbf{62.6$\pm$1.7} & \textbf{27.3$\pm$1.2} \\
 &  & heads6 & \textbf{75.5$\pm$0.1} & \textbf{48.8$\pm$0.2} & 59.1$\pm$1.6 & \textbf{27.3$\pm$1.2} \\
\cmidrule(l){2-7}
 & \multirow{2}{*}{SL-GCN} & all16 & \textbf{90.0$\pm$0.1} & \textbf{78.4$\pm$0.3} & \textbf{47.9$\pm$2.6} & \textbf{8.6$\pm$3.7} \\
 &  & heads6 & 88.8$\pm$0.3 & 75.3$\pm$0.4 & 43.2$\pm$4.9 & 5.0$\pm$1.0 \\
\cmidrule(l){2-7}
 & \multirow{2}{*}{SHuBERT} & all16 & \textbf{81.7$\pm$0.2} & \textbf{64.9$\pm$0.2} & 59.9$\pm$1.3 & 18.3$\pm$4.2 \\
 &  & heads6 & 79.9$\pm$0.2 & 60.1$\pm$0.3 & \textbf{62.8$\pm$3.0} & \textbf{18.5$\pm$3.3} \\
\bottomrule
\end{tabular}%
}
\caption{Supervision head-set ablation across all three model families. Models are grouped by training set (PopSign and Sem-Lex blocks) and evaluated on two test sets: in-domain (ASL Test) and zero-shot (LSC Zero-Shot). Setting: full 16-head training (\texttt{all16}) vs.\ 6 handshape-relevant heads (\texttt{heads6}). Ph.: mean accuracy across the handshape phonological features; Hs.: expected handshape accuracy. Cells report mean $\pm$ standard deviation across seeds; \textbf{bold} denotes the higher value per paired ablation setting.}
\label{tab:heads_ablation}
\end{table}

\section{Ablation on Target Phonological Heads}
\label{app:ablation_target_heads}

To evaluate the impact of auxiliary multi-task supervision, we compare models trained to predict all 16 phonological features (denoted as \texttt{all16}) against those trained exclusively on the 6 handshape-related heads, i.e. the five handshape phonological features alongside the handshape class, (denoted as \texttt{heads6}), as introduced in Section~\ref{par:target_phonological_heads}. 

Table~\ref{tab:heads_ablation} details the empirical results, revealing a clear trade-off between in-domain performance and zero-shot transfer capability. For in-domain ASL evaluation, the full 16-head auxiliary supervision (\texttt{all16}) acts as a valuable multi-task regularizer for the more complex architectures (SL-GCN and SHuBERT), consistently yielding higher accuracy than \texttt{heads6}. 
However, in zero-shot LSC transfer, restricting supervision to \texttt{heads6} frequently results in competitive or even superior performance. For example, Sem-Lex-trained SHuBERT improves from $59.9\%$ to $62.8\%$ (Ph.) when discarding the auxiliary non-handshape heads. Similarly, the simpler MLP baseline slightly benefits from the focused \texttt{heads6} setting across both datasets. This suggests that while full multi-tasking helps robustly model in-domain ASL phonology, focusing solely on handshape variables can sometimes prevent over-specialization to ASL's specific feature distributions, aiding cross-lingual generalization.
\newpage

\end{document}